\documentclass[journal]{IEEEtran}
\usepackage{graphicx}
\usepackage{algorithm} 
\usepackage{algorithmic} 
\usepackage{color}
\usepackage{amsmath}
\usepackage{CJK}
\usepackage{flushend}
\usepackage{amsfonts}
\usepackage{mathrsfs}
\usepackage{booktabs}
\usepackage{hyperref}
\usepackage{multirow}
\usepackage{hyperref}

\ifCLASSINFOpdf
\else
\fi

\begin{document}
\title{Context-Patch Face Hallucination Based on Thresholding Locality-constrained Representation and Reproducing Learning}


\author{Junjun~Jiang,~\IEEEmembership{Member,~IEEE,}
        Yi~Yu,
        Suhua~Tang,~\IEEEmembership{Member,~IEEE,}
        Jiayi Ma,~\IEEEmembership{Member,~IEEE,}
        Akiko Aizawa,
        and Kiyoharu Aizawa,~\IEEEmembership{Fellow,~IEEE}

\thanks{The research was supported by the National Natural Science Foundation of China under Grants 61501413, 61503288 and 61773295, and was also partially supported by JSPS KAKENHI Grant Number 16K16058.}
\IEEEcompsocitemizethanks{
\IEEEcompsocthanksitem J. Jiang is with the School of Computer Science and Technology, Harbin Institute of Technology, Harbin 150001, China, and is also with the Digital Content and Media Sciences Research Division, National Institute of Informatics, Tokyo 101-8430, Japan (junjun0595@163.com).
\IEEEcompsocthanksitem Y. Yu and A. Aizawa are with the Digital Content and Media Sciences Research Division, National Institute of Informatics, Tokyo 101-8430, Japan (\{yiyu, aizawa\}@nii.ac.jp).
\IEEEcompsocthanksitem S. Tang is with the Department of Communication Engineering and Informatics, The University of Electro-Communications, Tokyo 182-8585, Japan (shtang@uec.ac.jp).
\IEEEcompsocthanksitem J. Ma is with the Electronic Information School, Wuhan University, Wuhan 430072, China (jyma2010@gmail.com).
\IEEEcompsocthanksitem K. Aizawa is the Department of Information and Communication Engineering, The University of Tokyo, Tokyo 113-8654, Japan (e-mail: aizawa@hal.t.u-tokyo.ac.jp).
}
}


\maketitle

\begin{abstract}
Face hallucination is a technique that reconstruct high-resolution (HR) faces from low-resolution (LR) faces, by using the prior knowledge learned from HR/LR face pairs. Most state-of-the-arts leverage position-patch prior knowledge of human face to estimate the optimal representation coefficients for each image patch. However, they focus only the position information and usually ignore the context information of image patch. In addition, when they are confronted with misalignment or the Small Sample Size (SSS) problem, the hallucination performance is very poor. To this end, this study incorporates the contextual information of image patch and proposes a powerful and efficient context-patch based face hallucination approach, namely Thresholding Locality-constrained Representation and Reproducing learning (TLcR-RL). Under the context-patch based framework, we advance a thresholding based representation method to enhance the reconstruction accuracy and reduce the computational complexity. To further improve the performance of the proposed algorithm, we propose a promotion strategy called reproducing learning. By adding the estimated HR face to the training set, which can simulates the case that the HR version of the input LR face is present in the training set, thus iteratively enhancing the final hallucination result. Experiments demonstrate that the proposed TLcR-RL method achieves a substantial increase in the hallucinated results, both subjectively and objectively. Additionally, the proposed framework is more robust to face misalignment and the SSS problem, and its hallucinated HR face is still very good when the LR test face is from the real-world. \textcolor[rgb]{1.00,0.00,0.00}{The MATLAB source code is available at https://github.com/junjun-jiang/TLcR-RL}.
\end{abstract}


\IEEEpeerreviewmaketitle

\section{Introduction}
\label{sec:intro}
Face hallucination, which can be seen as a domain-specific super-resolution technology, is a technique to infer a High-Resolution (HR) face image, along with increasing the detailed face features, from low-resolution (LR) face images~\cite{Wang2013Survey}. It has numerous applications for face recognition~\cite{Zou2010TIP, shi2015from}, 3D face modeling~\cite{LiuJY2016TCYB}, criminal detection~\cite{wang2017bayesian, lin2005investigation}, and so on. From the pioneering work of~\cite{Baker2000, Liu2001}, many issues of face hallucination have been increasingly studied~\cite{Ma2015robust, LiuLC2018Quaternion}.
Generally speaking, these methods all try to explore the implicit or explicit transformation between the LR and HR spaces with an additional training set with LR and HR face image pairs. Most methods in the literature fall into two main categories: Statistical model based global face methods and patch prior based local face methods. \emph{A list of face hallucination resources collected by Jiang can be found at \cite{jiang2018arXiv}}.

Statistical model based global face methods leverage the face statistical models, such as PCA~\cite{Wang2005Eig}, locality preserving model~\cite{Park2007}, uniform space projection~\cite{Huang2010CCA, Anle2014face} and Nonnegative Matrix Factorization (NMF)~\cite{Yang2010TIP}, to model the face image and execute face hallucination globally. They can well maintain the global structure of human face. However, their results lack detailed local face features and suffer from ghosting artifacts.

Considering that the human face structure is a significant prior, many face hallucination methods try to exploit the prior knowledge present in smaller patches~\cite{Ma2010LSR, li2014face, Wang2014TCSVT, hui2017novel}. Among them, position-patch based methods have gained widespread attention in recent years. The common idea of these methods is to divide the global face into many small patches with predefined patch size and overlap, and use the training patches with the same position as the input one to construct the input patch. In this paper, our work is mainly concerned with this type of approach.

The Least Square Representation (LSR) method~\cite{hu2016serf} is one of {the representative position-patch based methods~\cite{Ma2010LSR}}. {To address the problem that the solution of LSR is unstable}, Sparse Representation (SR) based models have been developed by incorporating the sparsity regularization~\cite{Yang2010TIP, Jung2011SR, Gao2015Neuro, Jiang2017TCYB}. However, SR methods overemphasize sparsity and neglect local similarity among the training samples, which is essential for exploiting the intrinsic non-linear manifold of the training sample space~\cite{Yu2009, Wang2010LLC}. The approach of~\cite{Jiang2014LcRTMM} develops a Locality-constrained Representation (LcR) model which simultaneously {adds} the sparsity and locality constraints to the patch representation objective function, obtaining stable and reasonable representation coefficients. In order to alleviate the inconsistency of the LR and HR spaces, some works have been proposed to iteratively obtain the patch representation and perform neighbor embedding or learn the mapping in correlation spaces~\cite{Anle2014face, farrugia2015face, Jiang2016CSVT, shi2018hallucinating}. Based on LcR, recently, the low-rank and self-similarity priors are also introduced to regularize patch representation in~\cite{gao2016locality, Lu2017Access, jiang2017srlsp}. {In \cite{pei2018face}, Pei \emph{et al.} incorporated the gradient information of face image to further regularize the patch representation.} In addition {to face hallucination}, the LcR algorithm has been also used to deal with pose and illumination problems in face hallucination and synthesis~\cite{Ma2015robust, LiuLC2018Quaternion, LiuLC2017Robust}.

\begin{figure}[t]
\centering
\centerline{\includegraphics[width=7.35cm]{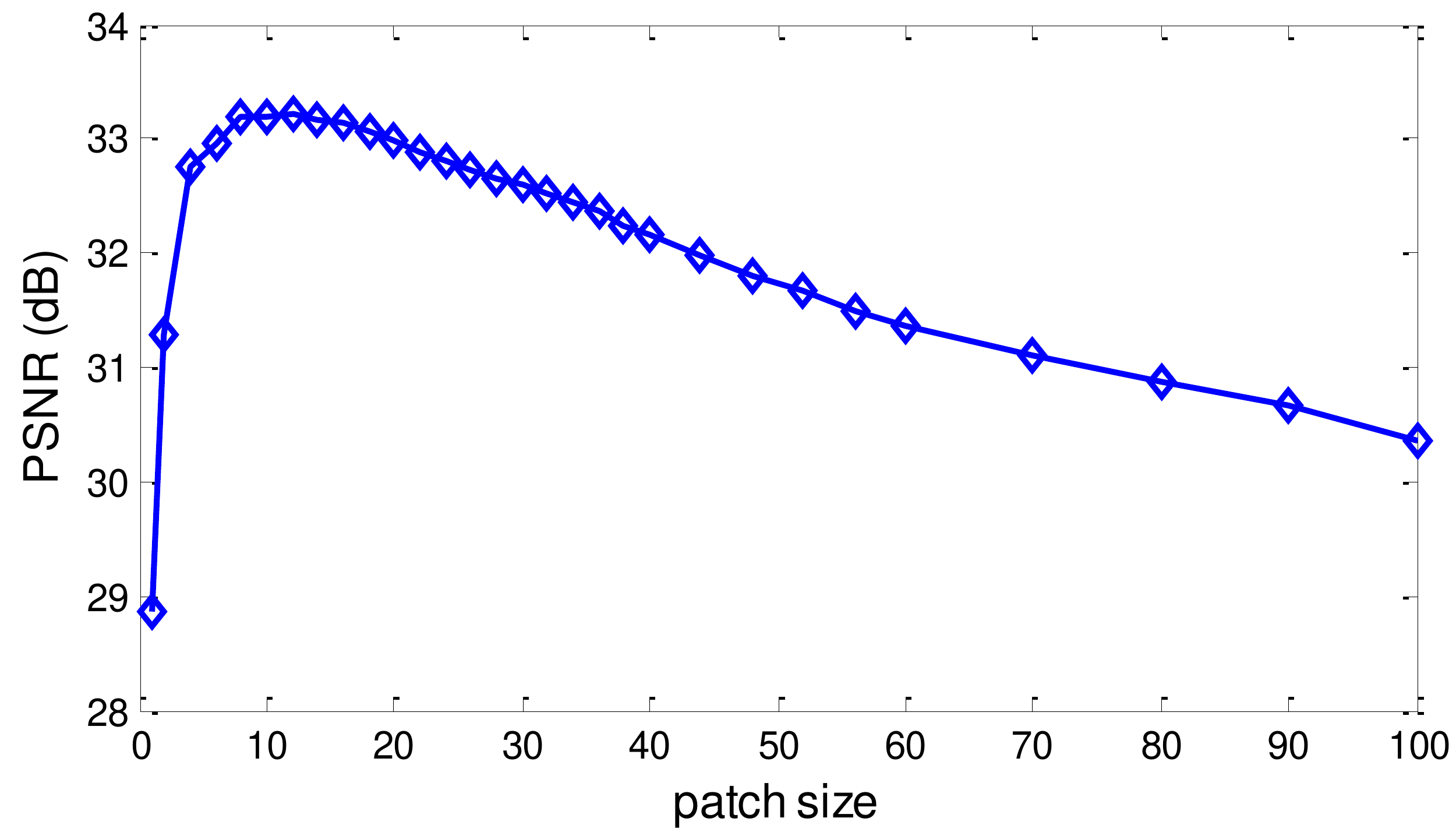}}
\vspace{-0.20cm}
\caption{Influence of the patch size over the position-patch based method~\cite{Jiang2014LcRTMM}. Simply enlarging the patch size does not {invariably} bring performance improvement. It reaches the best performance when the patch size is set to 12 in the FEI database \cite{FEI} with training sample size of 360.}
\label{fig:pppatchsize}
\end{figure}

However, aforementioned local patch treatments mainly focus on the small patches and do not take into account the global nature, which has been verified to be beneficial to image description, image denoising and retrieval tasks~\cite{SunJian2010CVPR, Levin2011CVPR,romano2016patch}. To model the global nature of local patch based methods, the most direct way to incorporating the contextual information is to extend the patch as discussed in~\cite{hao2015unified, chen2016face}. {The most extreme situation is to treat the entire face as a whole, using a global face-based approach. Another possible solution is to introduce a global reconstruction constraint in the image patch based method \cite{shi2014global, shi2015kernel, shi2018hallucinating}.} However, when the training sample size is fixed, it will become much more {formidable} to reconstruct a large patch or infer the global face image. In other words, because the training sample size should grow exponentially with the size of the image patch, it becomes impractical to present a too large image patch~\cite{romano2016patch, shi2014global}. This point is illustrated by Fig. \ref{fig:pppatchsize} (to avoid the effect of overlap level under different patch sizes, we set the overlap pixel to the half of patch size).
More recently, to reconstruct the latent HR image locally while thinking globally, DNNs, especially CNNs, have been applied to construct the mapping relationship between the LR images and their HR counterparts and shown strong learning capability and accurate prediction of HR images \cite{dong2016image, liu2016robust}. 
For example, Dong \emph{et al.}~\cite{dong2016image} developed a general image super-resolution method based on SRCNN. This is the very first attempt to use deep learning tools for image super-resolution reconstruction.
The approach of Liu \emph{et al.} \cite{liu2016robust} proposes to introduce the domain expertise to design a Sparse Coding based Network (SCN). Recently, R-DGN~\cite{yu2016ultra}, CBN~\cite{zhu2016deep}, LCGE~\cite{song2017learning}, Attention-FH \cite{Cao2017CVPR}, FSRNet \cite{CT-FSRNet-2018}, and \cite{Yu2018Imagining} are the most competitive approaches for face hallucination. They unitized very deep networks to model the relationship between the LR images and their HR correspondings, and verified that deeper networks can produce better results due to the large receptive field, which means considering more contextual information, \emph{i.e.}, very large image regions.

\begin{figure}[t]
\centering
\centerline{\includegraphics[width=8.95cm]{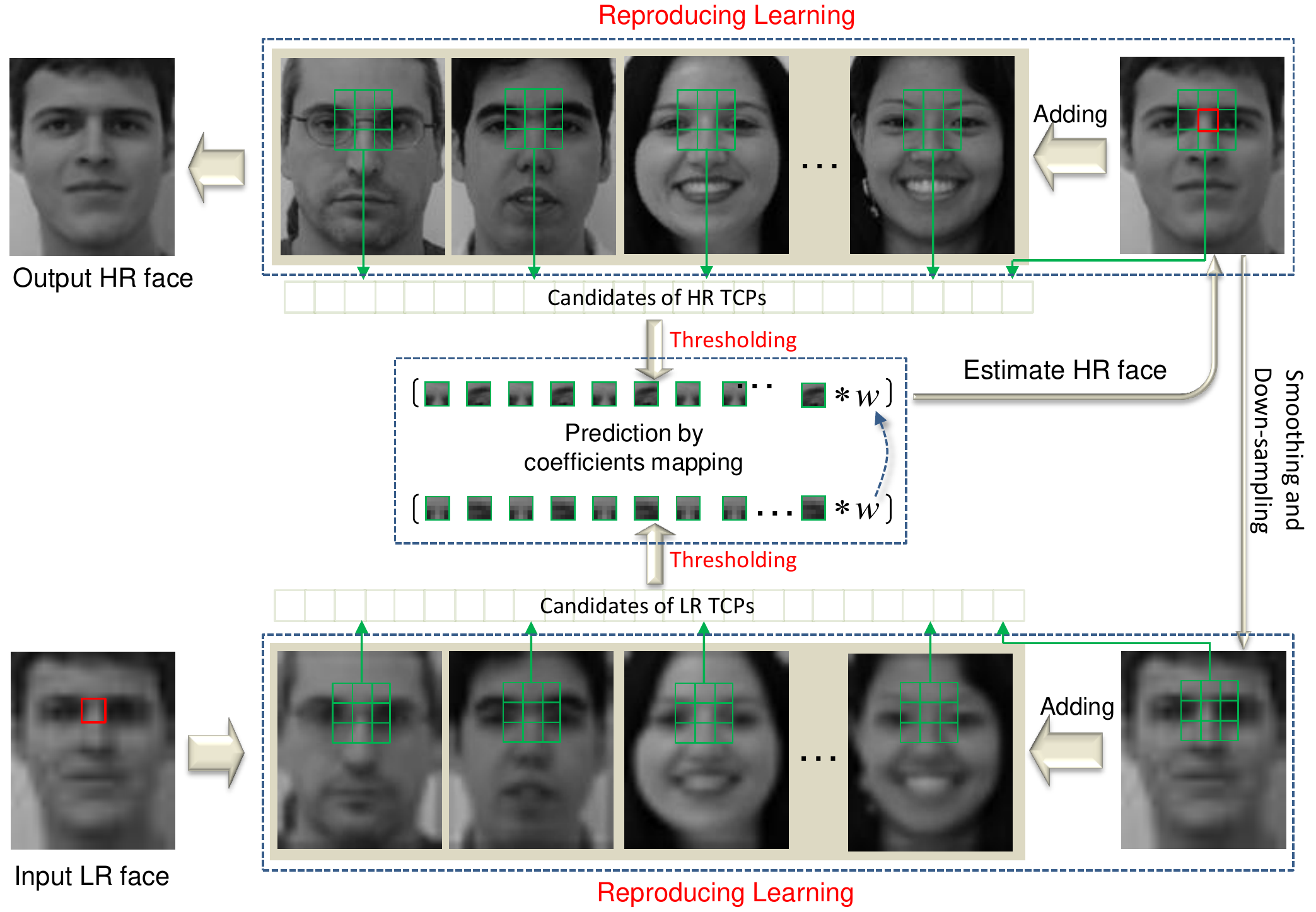}}
\vspace{-0.20cm}
\caption{Flow diagram of our proposed TLcR-RL based context-patch face hallucination framework. The face images marked with gray background are the HR and LR training sample pairs.}
\label{fig:Schematic_diagram}
\end{figure}

\subsection{Motivation and Contributions}
To utilize the contextual information while without enlarging the patch size, this paper proposes to simultaneously consider all the patches in a large window centred at the observation patch that named \emph{context-patch} and develop a context-patch based face hallucination framework. It inherits the merits of predicting with small local patches, while having the benefits of working with large patches (large receptive field). Based on Thresholding LcR (TLcR), the stability of representation and reconstruction accuracy can be improved. {Observing that the reconstruction performance will be improved if there are some similar samples in the training set}, we {further} advance an enhancement scheme via Reproducing Learning (RL), {which puts reconstructed HR samples back to the training set and makes it easier to reconstruction the input image.} As shown in Fig.~\ref{fig:Schematic_diagram}, we illustrate the framework of our proposed context-patch based face hallucination algorithm. For a testing patch, which is marked by red box, on the input LR face image, we first extract the LR Context-Patches (TCPs), which are marked by green boxes, from the LR training set. Then, we calculate the distances between the input LR patch and TCPs, and choose the $K$ nearest neighbor patches to reconstruct the input LR patch. Lastly, the {output} HR patch can be prediced by combining the corresponding HR TCPs with the representation coefficient $\textbf{w}$ obtained in the LR training set. To promote the performance, we add back the hallucinated HR image to the training set, which can simulates the case that the HR version of the input LR face is present in the training set, and repeat the \emph{thresholding-representation-prediction} steps to iteratively enhance the final hallucination result. In summary, the main contributions of this study are threefold.
\begin{itemize}
  \item {We introduce the concept of context-patch to expand the receptive field of the patch representation model. It not only inherits the merits of predicting with small patches but also has the benefits {of} working with large patches. In addition, we combine the low-level pixel values and high-level position information to represent the image patch, thus further exploiting the contextual information.}
  \item {We develop a novel and robust image patch reconstruction method based on thresholding locality-constrained representation. It is inherited from the LcR method~\cite{Jiang2014LcRTMM}, but has the advantages of accurate patch representation and low computational complexity.}
  \item We propose a face hallucination improvement strategy via reproducing learning. The estimated HR face image is iteratively reconstructed with a reproduced training set through adding the hallucinated HR image and its degenerated version to the training set. Experiments demonstrate its superiority some state-of-the-arts in term of both objective assessment and visual quality, especially when confronted with misalignment or the SSS problem.
\end{itemize}

{The} research reported in this paper is an extension of our preliminary work~\cite{Jiang2017ICME}. We highlight the significant differences between this research and~\cite{Jiang2017ICME} as follows: (i) to exploit much more contextual information of the patch images, we extent the pixel intensity based representation to the combination of low-level pixel values and high-level position prior (can be seen as the contextual information). (ii) the approach of ~\cite{Jiang2017ICME} focuses only on controllable conditions. However, this research extends the application of TLcR-RL algorithm from controllable conditions to more {intricate} conditions, including both very limited training sample size (the SSS problem) and real-world image reconstruction. (iii) this research gives deep analysis on the motivations and advantages of introducing the contextual information, thresholding strategy, and reproducing learning, leading to a better understanding of why and how our method works. 

\subsection{Organization of This Paper}
\label{Organization}
The rest parts of this study is organized as follows. In Section \ref{sec:formulation}, we give some notations and present the formulation of position-patch based methods. Section~\ref{sec:proposedmethod} presents the details of the TLcR-RL based face hallucination method followed by the improvement strategies of thresholding based patch representation and reproducing learning based iterative estimation. In Section~\ref{sec:experiments}, we report experimental evaluations of the context-patch based face hallucination framework and compare with some competitive algorithms. Finally, we conclude this paper and present the possible future work in the last section.

\section{Preliminaries}
\label{sec:formulation}

In this paper, HR training set (consists of all HR training samples) is denoted as $\mathscr{Y}_H=\{ \textbf{Y}_H^m\} _{m = 1}^M$ and their LR counterpart (consists of all LR training samples) is denoted as $\mathscr{Y}_L=\{ \textbf{Y}_L^m\} _{m = 1}^M$, where $\textbf{Y}_H^m$ ($\textbf{Y}_L^m$) denotes the HR (LR) training samples with index $m$, and $M$ is the size of training sample. 

For position-patch methods, HR training images, LR training images, and the observed LR image are all divided into image patches according to the position information, $\{\textbf{y}_H^m(p)\}_p$, $\{\textbf{y}_L^m(p)\}_p$ and $\{\textbf{x}_L(p)\}_p$, respectively, $p$ is the position index. Given the LR testing patch ${\textbf{x}_L}(p)$, the position-patch based method tries to utilize different constraints to regularize the representation coefficients $\textbf{w}_L(p)$ in the LR training space:
\begin{equation}\label{eq:position}
J(\textbf{w}_L(p)) = \left\| {{\textbf{x}_L}(p) - \sum\limits_{m = 1}^M {\textbf{y}_L^m(p){w_L^m}(p)} } \right\|_2^2 + \tau \Omega (\textbf{w}_L(p)),
\end{equation}
where $\textbf{w}(p)$ refers to the prior about the reconstruction weights. $\tau$ is a locality regularization parameter used to balance the contributions between the reconstruction errors and prior knowledge, \emph{i.e.}, the closeness to the LR testing patch and the desired properties of the representation coefficients.

Based on the {Locally Linear Embedding (LLE)}~\cite{Roweis2000} manifold learning assumption that HR and LR samples share similar local geometrical structure~\cite{Chang2004NE}, which is characterized by the reconstruction coefficients in a neighborhood, these patch representation methods directly transform the LR representation coefficients to the HR space, \emph{i.e.}, let $\textbf{w}_H(p)=\textbf{w}_L(p)$:
\begin{equation}\label{eq:HRpredict}
{\textbf{x}_H}(p) = \sum\limits_{m = 1}^M {\textbf{y}_H^m(p){w_L^m}(p)}.
\end{equation}

Upon acquiring all the estimated HR face image patches ${\{ {\textbf{x}_H}(p)\} _p}$, the target HR face image can be calculated by placing all the estimated HR patches into original position and averaging each pixel from different reconstruction patches. For the simplicity of notation, we remove the term ``$(p)$'' in the following.

\section{Context-Patch Based Face Hallucination}
\label{sec:proposedmethod}
This section presents the proposed context-patch based face hallucination approach. First, we give the formulation of context-patch locality constrained representations. Then, we present the proposed thresholding approach to locality constrained presentation and reproducing learning strategy which {aims} to improve the reconstruction performance. Finally, we summarize the details of {the} face hallucination algorithm.

\subsection{Context-Patch Representation}
To construction the input LR face patch with the position of $p$, we use all the context-patches around position $p$ to obtain its reconstruction coefficients through the following objective function:
\begin{equation}\label{eq:context}
J(\textbf{w}_{L}) = \left\| {{\textbf{x}_L} - \sum\limits_{n = 1}^N {\textbf{y}_{LTCP}^n{w_L^n}} } \right\|_2^2 + \tau \Omega (\textbf{w}_{L}),
\end{equation}
where $N$ is the number of LR TCPs. The value of $N$ is determined by the window size ($w$), patch size($p$) and step size ($s$):
\begin{equation}\label{eq:number}
N = M {\left( {1 + \frac{{w - p}}{{s}}} \right)^2}.
\end{equation}
In this paper, we fix $s$ to 2. The values of $w$ and $p$ are set experimentally.

We denote by $\textbf{Y}_{LTCP}$ the LR TCPs matrix,
\begin{equation}\label{eq:nonumber1}
{\textbf{Y}_{LTCP}} = [{\rm{\textbf{y}}}_{LTCP}^1,\textbf{y}_{LTCP}^2,...,\textbf{y}_{LTCP}^N],\nonumber
\end{equation}
and ${\textbf{w}_{L}}$ the representation vector,
\begin{equation}\label{eq:nonumber2}
{\textbf{w}_{L}} = {[w^1_{L},w^2_{L},...,w^N_{L}]^T},\nonumber
\end{equation}
where $\textbf{y}^{i}_{LTCP}$ is the LR TCP and $w^i_{L}$ is its representation coefficient, $1 \le i \le N$.

%

With these definitions, we can rewrite the Eq.~(\ref{eq:context}) as follows,
\begin{equation}\label{eq:contextMatrix}
\begin{array}{l}
J({\textbf{w}_{L}}) = \left\| {{\textbf{x}_L} - \textbf{Y}_{LTCP}{\textbf{w}_{L}}} \right\|_2^2 + \tau \Omega ({\textbf{w}_{L}}).
\end{array}
\end{equation}


Similar to \cite{Jiang2014LcRTMM}, in this paper we employ the locality prior to regularize the representation of the input LR patch,
\begin{equation}\label{eq:contextMatrixLcR}
\begin{array}{l}
 J({\textbf{w}_{L}}) = \left\| {{\textbf{x}_L} - \textbf{Y}_{LTCP}{\textbf{w}_{L}}} \right\|_2^2 + \tau \left\| {\textbf{d} \odot {\textbf{w}_{L}}} \right\|_2^2 \\
{\kern 40pt} \text{s.t.}{\kern 1pt} {\kern 1pt} {\kern 1pt} {\kern 1pt} {\kern 1pt} \sum\limits_{i = 1}^N {w^i_{L}} {\kern 1pt} {\kern 1pt}  = 1 \\
 \end{array}
\end{equation}
where ``$ \odot $'' denotes the element by element product of two vectors, $\textbf{d}$ is a $N \times 1$ vector, $\textbf{d} = {[{d_1},{d_2},...,{d_N}]^T}$, with
\begin{equation}\label{eq:distance}
{d_n} = {\left\| {{\textbf{x}_L} - \textbf{y}_{LTCP}^{n}} \right\|_2}, 1 \le n \le N.
\end{equation}
In Eq.~(\ref{eq:contextMatrixLcR}), the sum-to-one constraint $\sum\limits_{n = 1}^N {w^n_{L}} {\kern 1pt} {\kern 1pt}  = 1$ is introduced to ensure that the reconstruction result is physically understandable.

By incorporating the {locality-constraint}, {different LR TCPs will receive different penalties (or freedom)}. Specifically, those patches {different} from the LR testing one will be heavily penalized and have relatively small representation coefficient, while those patches similar to the LR testing one will receive relatively large representation coefficient, which is consistent with the intuitive understanding.

\subsection{Thresholding Locality-constrained Representation (TLcR)}
Compared with traditional position-patch based method, our context-patch based method can incorporate much more patches ($N/M$ times) to construct the input patch. Thus, the representation ability of our method can be greatly promoted. However, this will also lead to two other problems: firstly, the multiplied increase in the training sample will lead to a rapid increase in computational complexity; secondly, many dissimilar image patches may be introduced to the training set, which will exacerbate the uncertainty. {This is mainly because that when the number of training samples is far more than the dimension of LR patch, the patch representation problem is seriously ill-posed and they are many solutions.} By selecting as few atoms (samples) as possible, it can expect to promote the patch representation stability as well as the reconstruction accuracy. The philosophy behind is same to the sparse representation and compressive sensing theory~\cite{Candes2006}.


Based on the above observations, this paper proposes an effective and efficient image patch representation algorithm based on a thresholding strategy, which selects the $K$ most similar training pathes to construct the input LR face image patch and sets the representation coefficients of other samples to zeros. Therefore, we can expect to greatly reduce the computational complexity and obviously improve the performance of the algorithm. In particular, we impose an additional regularization term to Eq. (\ref{eq:contextMatrixLcR}) to select the $K$ most similar LR TCPs pathes and discard the rest,
\begin{equation}\label{eq:TLcR1}
\begin{array}{l}
 J({\textbf{w}_{L}}) = \left\| {{\textbf{x}_L} - \textbf{Y}_{LTCP}{\textbf{w}_{L}}} \right\|_2^2  + \tau \left\| {\textbf{d} \odot {\textbf{w}_{L}}} \right\|_2^2\\
 \text{s.t.}{\kern 1pt} {\kern 1pt} {\kern 1pt} {\kern 1pt} {\kern 1pt} \sum\limits_{i = 1}^N {w_L^{i}} {\kern 1pt} {\kern 1pt}  = 1{\kern 1pt} {\kern 1pt} {\kern 1pt} {\kern 1pt} {\kern 1pt} {\kern 1pt} {\rm{and}}{\kern 1pt} {\kern 1pt} {\kern 1pt} {\kern 1pt} w_L^{k} = 0{\kern 1pt} {\kern 1pt} {\kern 1pt} {\rm{if}}{\kern 1pt} {\kern 1pt} {\kern 1pt} k \notin {\mathbb{C}_K}\left( {{\textbf{x}_L}} \right), \\
 \end{array}
\end{equation}
where ${\mathbb{C}_K}\left( {{\textbf{x}_L}} \right)$ represents the indices of $K$ nearest neighbor (KNN) of ${\textbf{x}_L}$ in $\textbf{Y}_{LTCP}$. By incorporating the addtional constraint, the coefficient is zero if $\textbf{y}_{LTCP}^{i}$ is not in the set of KNN. Therefore, we directly use the KNN to construct ${\textbf{x}_L}$,
\begin{equation}\label{eq:TLcR2}
\begin{array}{l}
 J({\textbf{w}_{L}^K}) = \left\| {{\textbf{x}_L} - \textbf{Y}_{LTCP}^K{\textbf{w}_L^{K}}} \right\|_2^2 +\tau \left\| {{\textbf{d}^K} \odot {\textbf{w}_L^K}} \right\|_2^2 \\
{\kern 40pt} \text{s.t.}{\kern 1pt} {\kern 1pt} {\kern 1pt} {\kern 1pt} {\kern 1pt} \sum\limits_{k \in {\mathbb{C}_K}\left( {{\textbf{x}_L}} \right)} {w^k_{L}}  = 1, \\
 \end{array}
\end{equation}
where $\textbf{Y}_{LTCP}^{K} = {\left\{ {\textbf{y}_{LTCP}^{k}} \right\}_{k \in {\mathbb{C}_K}\left( {{\textbf{x}_L}} \right)}}$ and ${\textbf{d}^K} = {\left\{ {d_k} \right\}_{k \in {\mathbb{C}_K}\left( {{\textbf{x}_L}} \right)}}$ are the $K$ nearest LR TCPs and the distances to these LR TCPs, respectively, ${\textbf{w}_L^{K}}={\left\{ {w_{L}^{k}} \right\}_{k \in {\mathbb{C}_K}\left( {{\textbf{x}_L}} \right)}}$ represents the representation coefficients of the $K$ nearest LR TCPs. Eq.~(\ref{eq:TLcR2}) is a convex quadratic problem and can be solved by an analytic optimal solution,
\begin{equation}\label{eq:TLcROptimization}
{\textbf{w}_{L}^K} = (\textbf{G}{^T}\textbf{G} + \tau {\textbf{D}^2})\backslash ones(K,1),
\end{equation}
where $\textbf{G} = {\textbf{x}_L} \cdot ones{(K,1)^T} - \textbf{Y}_{LTCP}^{K}$, $\textbf{D}$ is a $K \times K$ diagonal matrix with ${\textbf{D}_{kk}} = d_k, {k \in {\mathbb{C}_K}\left( {{\textbf{x}_L}} \right)}$, and $ones(K,1)$ is a $K \times 1$ column vector whose elements are all ones. The final optimal representation coefficients are obtained by rescaling ${\textbf{w}_{L}^K}$ to satisfy $\sum\limits_{k \in {\mathbb{C}_K}\left( {{\textbf{x}_L}} \right)} {w_L^{k}}  = 1$.

Acquiring the optimal representation coefficients ${\textbf{w}_{L}^K}$, the target HR patch ${y_L}$ can be predicted by:
\begin{equation}
{\textbf{y}_L} = \textbf{Y}_{HTCP}^{K}{\textbf{w}_L^{K}}.
\label{eq:HRprediction}
\end{equation}
where $\textbf{Y}_{HTCP}^{K} = {\left\{ {\textbf{y}_{HTCP}^{k}} \right\}_{k \in {\mathbb{C}_K}\left( {{\textbf{x}_L}} \right)}}$ denotes the corresponding $K$ nearest HR TCPs.

\begin{figure}[t]
\centering
\centerline{\includegraphics[width=8.00cm]{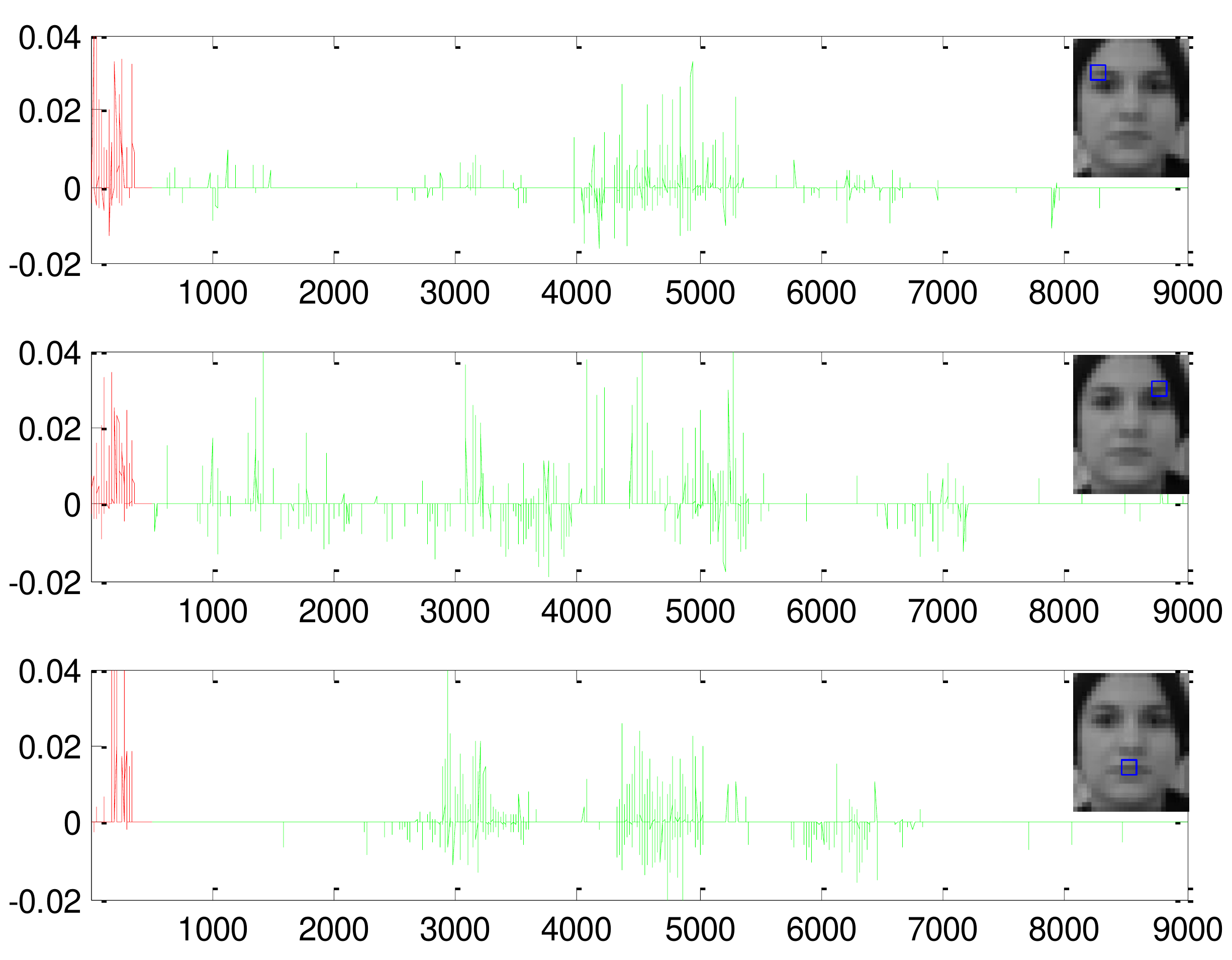}}
\vspace{-0.20cm}
\caption{Representation coefficients of some LR testing patches, which are marked by blue boxes. The red lines are the representation coefficients corresponding the position patches, while the green lines are representation coefficients corresponding the surrounding contextual patches. Here, the surrounding contextual patches denote all the TCPs except for the position-patches.}
\label{fig:threeRepresentations2}
\end{figure}

\begin{figure}[t]
\centering
\centerline{\includegraphics[width=7.50cm]{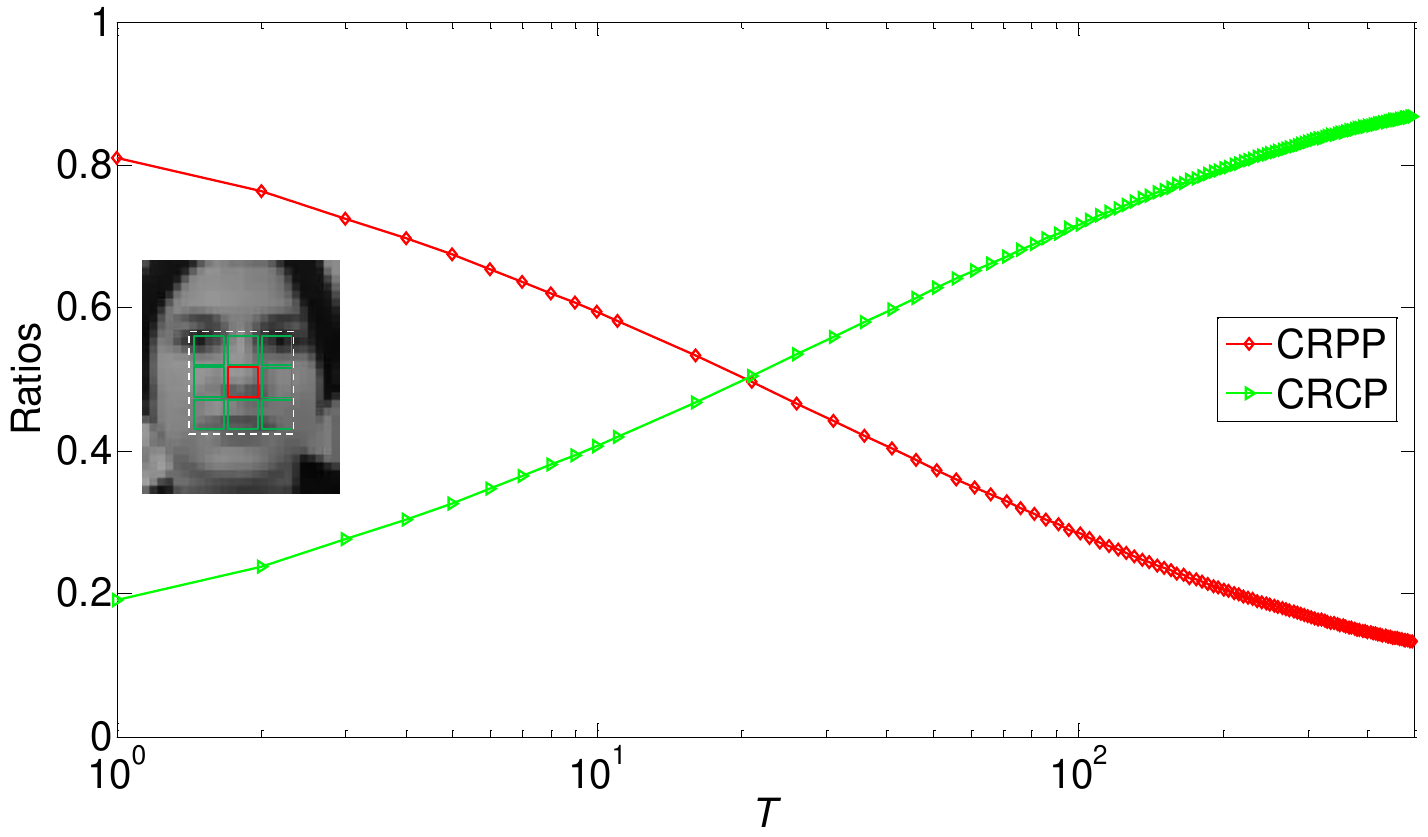}}
\vspace{-0.20cm}
\caption{Contribution analysis of position patches and surrounding contextual patches. $CRPP$: Contribution Ratio of Position-Patches, $CRCP$: Contribution Ratio of surrounding Contextual Patches. We plot the $CRPP$ and $CRCP$ according to different values of $T$.}
\label{fig:ratios}
\end{figure}

Through finding KNN, it transforms the large linear system to a small one, reducing the computation complexity of the linear system from $O(p^2KN+p^2K^3)$ to $O(p^2N^3)$, where $O(p^2KN)$ represent the additional complexity of $K-NN$ search in our method. We test the performance of with and without thresholding, we learn that the proposed thresholding scheme achieves 40 times faster than the original LcR method. In particular, the average running time of TLcR for one testing image is 13.8 seconds, while the LcR method will cost around 10 minutes.

To demonstrate the influence of position-patches and context-patches, in Fig.~\ref{fig:threeRepresentations2} we show where these patches that have non-zero coefficients come from.
Clearly, we can see that these position-patches and surrounding contextual patches simultaneously contribute to the reconstruction of the testing patch. In addition, we also qualitatively test the contribution ratio of these two kinds of TCPs. Therefore, we define a metric called contribution ratio of position-patches ($CRPP$ for short):
\begin{equation}
CRPP(T) = \frac{{\# \left\{ {in{d_T} \cap ind{_{pp}}} \right\}}}{T},
\label{eq:ratios}
\end{equation}
where ${ind{_T}}$ denotes the indices of the $T$ most {significant} patches (the larger the values in the representation vector $w$ are, the more {significant} the corresponding patches will be), ${ind{_{pp}}}$ denotes the indices of the position patches, and ${\# \left\{  \cdot  \right\}}$ represents the cardinality of a set. Therefore, the contribution ratio of surrounding contextual patches ($CRCP$ for shot) is $CRCP(T) = 1-CRPP(T)$. In Fig.~\ref{fig:ratios}, we plot the $CRPP$ and $CRCP$ according to $T$. We find that (i) when $T=1$, $CRPP(1)$ is 81\% and $CRCP(1)$ is 19\%; (ii) when $T=21$, $CRPP(21) \approx CRCP(21)\approx50\%$; {(iii) when $T=360$, $CRPP(360)$ is 15\% and $CRCP(360)$ is 85\%;} (iV) when $T=500$, $CRPP(500)$ is 12\% and $CRCP(500)$ is 88\%. It demonstrates that in addition to the position patches, the contextual surrounding patches are also very {significant} for the patch representation, and {not all significant patch necessarily come from the position-patches}.


\subsection{Reproducing Learning}
The performance of learning based face hallucination methods (or general image super-resolution methods) usually {depends} on the distribution similarity between the training and testings samples, \emph{i.e.}, the similarity between training and testing faces. If the HR face is preset in the training set, the reconstruction result is excellent. In contrast, when the original HR face of the input LR face is not in the training set and the input is dissimilar to samples in the training set, the performance of face hallucination algorithm will be very poor~\cite{wang2013transductive}.




Inspired by the above observations, in this paper we propose to add the hallucinated HR face image to the training set, which can simulates the case that the HR version of the input LR face is present in the training set, and then perform TLcR based face hallucination with this newly generated training set (as show in Fig.~\ref{fig:Schematic_diagram}). By reproducing learning, it will obtain 0.40 dB gain in term of Peak Signal-to-Noise Ratio (PSNR)  and 0.0043 gain in term of Structural SIMilarity (SSIM)~\cite{Wang2004SSIM} (please refer to the experimental section).

{In order to theoretically and technically explain the effectiveness of the proposed reproducing learning strategy, we give the following analysis. As a patch-based face hallucination approach, the hallucinated HR face is not {necessarily} the linear combination of training samples. By decomposing the entire face into smaller patches, the number of training patches will be greater than the patch size, the dictionary will become over-complete and it can reconstruct one patch with any content. That¡¯s to say, in the absence of any other constraints, these patch-based methods can reconstruct any image which never appears in the database, e.g., a cat or a dog image, and thus introduce extra information to the training data. When the size of training set is smaller than the dimensionality of the face image, the face image to be reconstructed may not lay at the space spanned by the training samples. Therefore, if we put reconstructed HR samples (by a patch-based method) back to the training set, we can actually introduce some additional information in the sense {that} we can add an image that cannot be linear combination with the original training samples.}

{In addition, we also explain the effectiveness of the proposed reproduce learning strategy from the perspective of dictionary learning. As reported in the literature \cite{aharon2006rm, Yang2010TIP}, how to learn a representative dictionary is crucial in the image reconstruction and analysis problems. Sparse coding based dictionary learning method is the most successful dictionary learning technique \cite{Yang2010TIP, LiuXM2016Compressive, LiuXM2017Sparsity}, which aims at generating an over-complete dictionary with atoms that are linear combination with the training samples. Although the learned dictionary does not introduce additional information, it has a stronger reconstruction capabilities than the non-extended training set. As for our proposed method, putting the reconstructed HR samples back to the training set can be seen as generating a much more adaptive dictionary to the observed image. Thus, better reconstruction performance can be expected.}

\begin{algorithm}[t]
\caption{\bfseries{Face Hallucination Based on TLcR-RL.}}
\begin{algorithmic}[1]
\label{algorithm1}
\STATE {\bfseries{Input}}: HR and LR training face pairs, $\mathscr{Y}_H=\{ \textbf{Y}_H^m\} _{m = 1}^M$ and $\mathscr{Y}_L=\{ \textbf{Y}_L^m\} _{m = 1}^M$, and an LR input $\textbf{X}_L$.
\STATE Divide the HR and LR training faces and the input LR face into small overlapping patches.
\FOR{each LR testing patch ${\textbf{x}_L}$}
\STATE Obtain the distance between ${\textbf{x}_L}$ and all the $N$ LR TCPs according to Eq.~(\ref{eq:distance}).
\STATE Select $K$ most similar LR TCPs of ${\textbf{x}_L}$.
\STATE Compute the optimal representation coefficients according to Eq.~(\ref{eq:TLcROptimization}).
\STATE Construct the HR patch according to Eq.~(\ref{eq:HRprediction}).
\ENDFOR
\STATE Restore the HR face image by placing all predicted HR patches according to their positions and averaging each pixel from different reconstruction patches.
\STATE Reproduce a novel training set by adding the estimated HR image and its degenerated LR image to the original training set.
\STATE Repeat Step 2-Step 10 until the iteration number is reached.
\STATE {\bfseries{Output}}: HR hallucinated  face image $\textbf{X}_{H}$.
\end{algorithmic}
\end{algorithm}


\subsection{The Overall Algorithm}
\begin{figure}[t]
\centering
\centerline{\includegraphics[width=8.75cm]{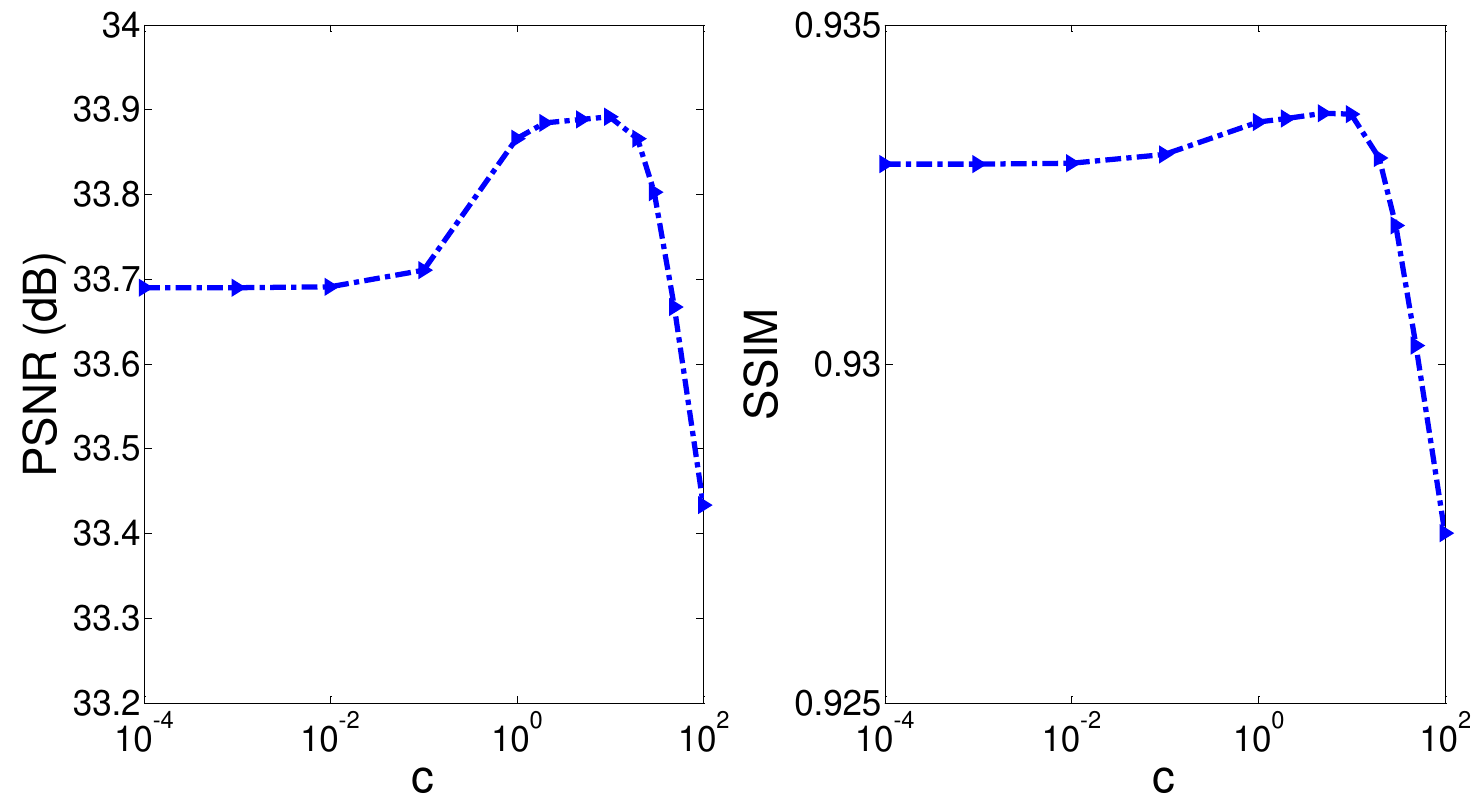}}
\vspace{-0.20cm}
\caption{Objective results according to different values of $c$. The performance will be stable when the value of $c$ is between 1 and 20.}
\label{fig:paraC}
\end{figure}

The complete face hallucination framework of our proposed TLcR-RL model is summarized as Algorithm \ref{algorithm1}. It should be noted that we extract the LR patch features by mean-removed pixel values. Moreover, in order to incorporate much more contextual information, we additionally incorporate the position information, the vertical and horizontal coordinates of one patch, to the feature representation. This can be seen as the high-level information and has been successfully used in recovering the depth structure of human face~\cite{LiuJY2016TCYB}. Specifically, we leverage the relative coordinates to denote the position information.
We use $\textbf{x}_L = [\textbf{x}_L; {f} \cdot p_x; {f}\cdot p_y]$ and $\textbf{y}_L = [\textbf{y}_L; {f} \cdot p_x; {f}\cdot p_y]$ to denote the feature representations of input LR patch and LR training patches. Here, ${f}$ is the weight of the position information in the representation, {and} $p_x$ and $p_y$ are the vertical and horizontal coordinates of one patch. We experimentally set the value of ${f}$ to 10 in our experiments, which will produce the best performance as shown in Fig. \ref{fig:paraC}. For the HR images, we extract their high-frequency components, by subtracting the interpolated LR image, as the features. In the prediction stage, we add the estimated HR image into the Bicubic interpolated LR image. The aforementioned feature extraction and high-frequency prediction approach can improve the hallucinated results. In the experiments, we additionally found that when we use joint features of raw pixels and high-level patch position information (which can be seen as the contextual information of patches), e.g., simply combining them into a longer feature vector with a balancing scalar that controls the importance of the contextual information, the overall performance of our method will have about 0.20 dB improvement over the original low-level intensity based patch representation method. 

\section{Experiments}
\label{sec:experiments}
This section evaluates the effectiveness of our proposed TLcR-RL method to face hallucination. Through these experiments, we can expect to answer following questions:

\begin{itemize}
  \item {Is the introduced contextual information helpful for face hallucination}?
  \item How does TLcR-RL compare against state-of-the arts?
  \item How does the algorithm perform with different training sample {sizes}, and does it works well when confronted with the SSS problem?
  \item How robust is the algorithm to misalignment?
  \item Is the proposed method useful in real-world scenarios?
\end{itemize}




\subsection{Experimental Settings}
\label{sec:database}
In the experiments, the public FEI database~\cite{FEI} is used. The HR faces are obtained by cropping the face image in FEI to 120$\times$100 pixels.
Similar to \cite{Jiang2014LcRTMM}, 360 images are randomly selected as the training samples, while the rest 40 images are employed for algorithm testing. Thus, all the testing face images are absent in the training set. All face images are aligned in the FEI database. In practical, we can apply the automatic alignment methods and feature points matching methods \cite{Ma13b, Ma2018Guided, Ma2019IF} to preprocess face images. Similar to~\cite{Jiang2014LcRTMM, Jiang2014TIP, gao2016locality}, we obtain the LR images by a filter (4$\times$4 average smoothing) and 4$\times$ down-sampling.
To balance the computational complexity and hallucination performance, the patch size and overlap pixels of all patch based methods are set to 12$\times$12 pixels and 4 pixels, respectively, as in~\cite{Jiang2014LcRTMM, Jiang2014TIP, gao2016locality}.

\subsection{Model Analysis}
\label{sec:Parameterselection}
In the proposed method, there are four parameters, \emph{i.e.}, the balancing parameter $\tau$ and the thresholding number $K$ in the objective function, the window size $w$, and the iterations in reproducing learning. By analyzing the above parameters on the performance of the algorithm, we can learn that the effectiveness of our proposed locality constraint, hard thresholding scheme, context-patch information, as well as the reproducing learning, are all testified.

\begin{figure}[t]
\centering
\centerline{\includegraphics[width=8.5cm]{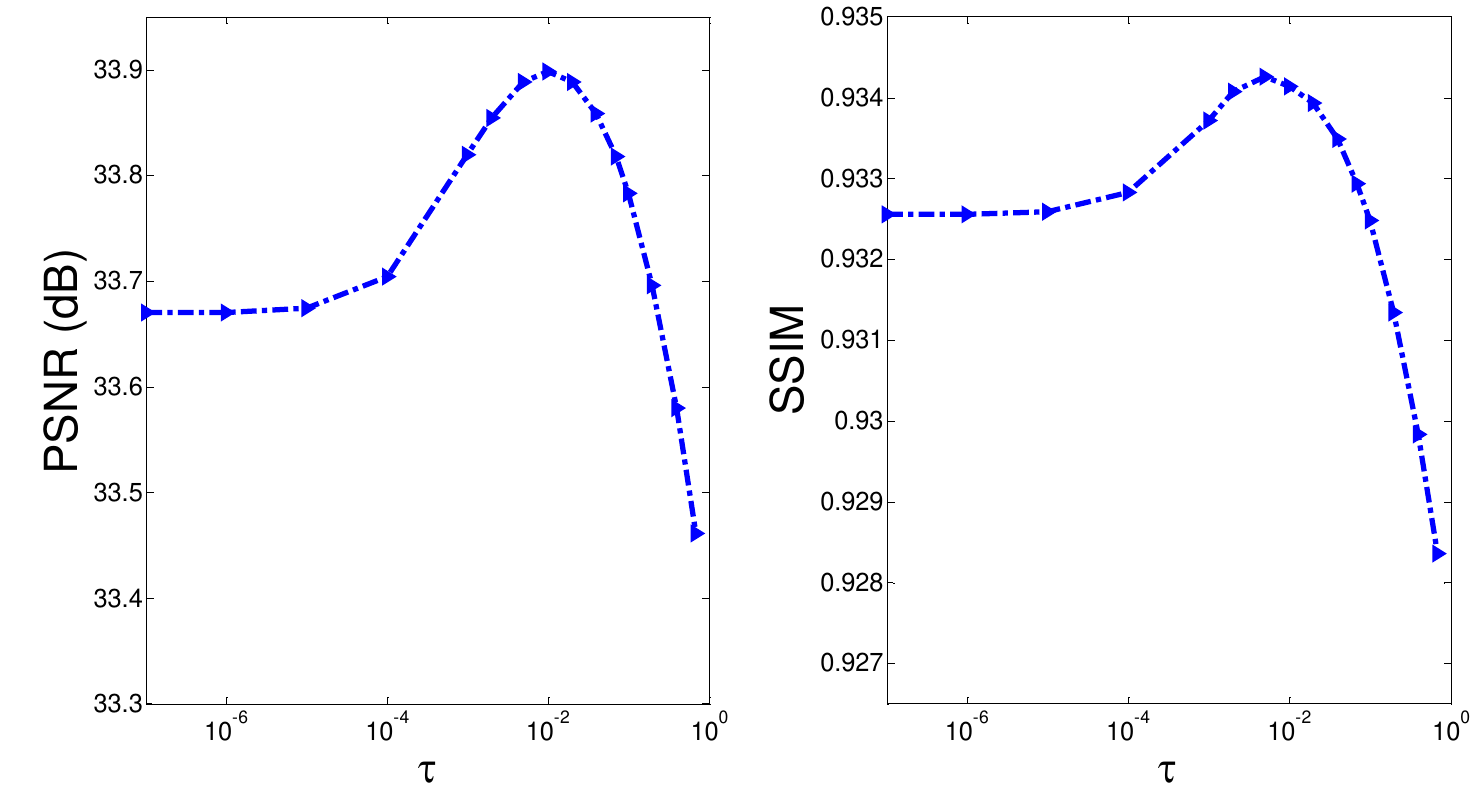}}
\vspace{-0.20cm}
\caption{Objective results in terms of average PSNR and SSIM of our proposed TLcR-RL based face hallucination method according to $\tau$.}
\label{fig:paraTau}
\end{figure}

\begin{figure}[t]
\centering
\centerline{\includegraphics[width=8.5cm]{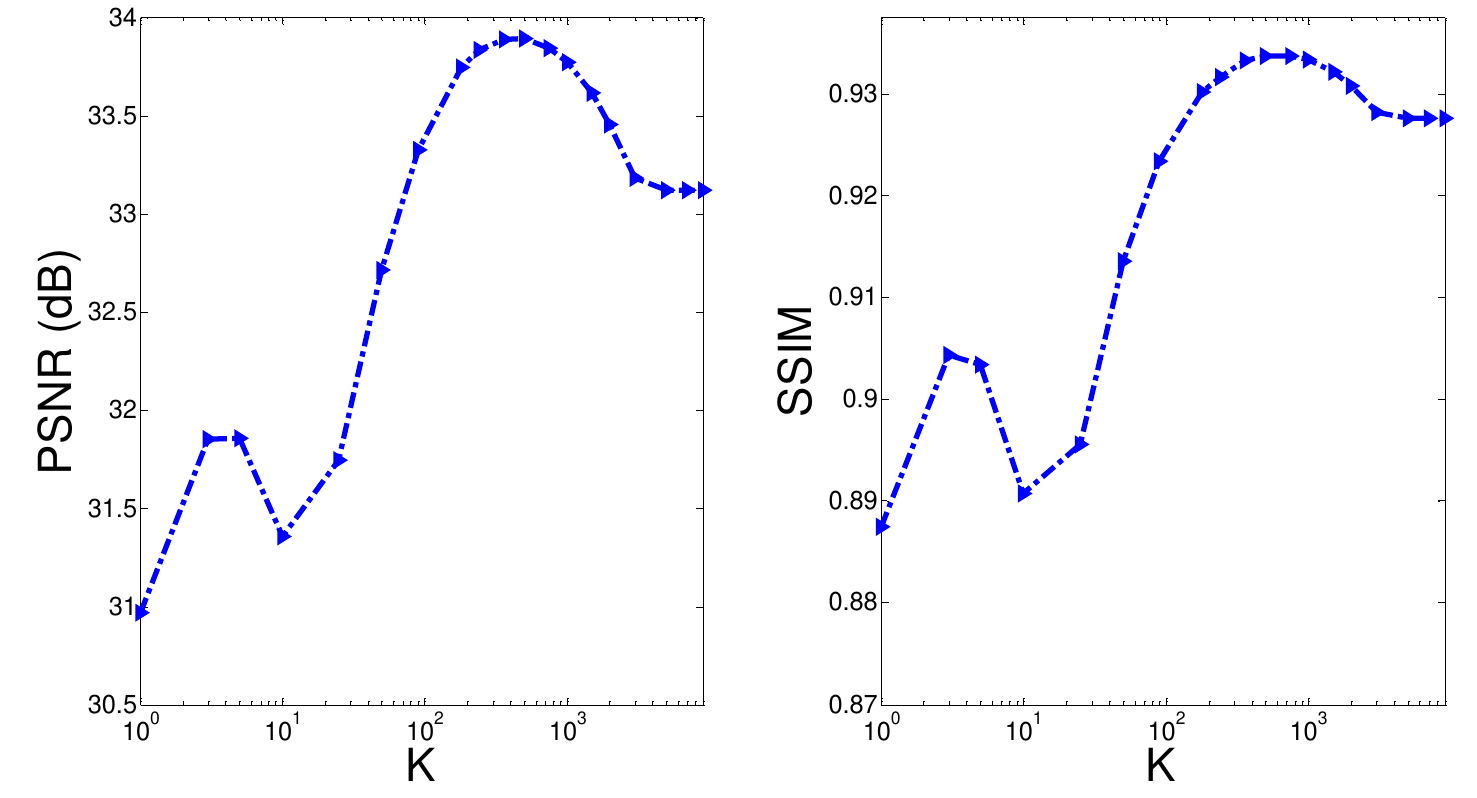}}
\vspace{-0.20cm}
\caption{Objective results of our proposed TLcR-RL based face hallucination method according to $K$. The decrease around 9 can be explained by the over-fitting on the input LR image patch~\cite{Bevilacqua2012}.}
\label{fig:paraK}
\end{figure}

\begin{figure}[t]
\centering
\centerline{\includegraphics[width=8.5cm]{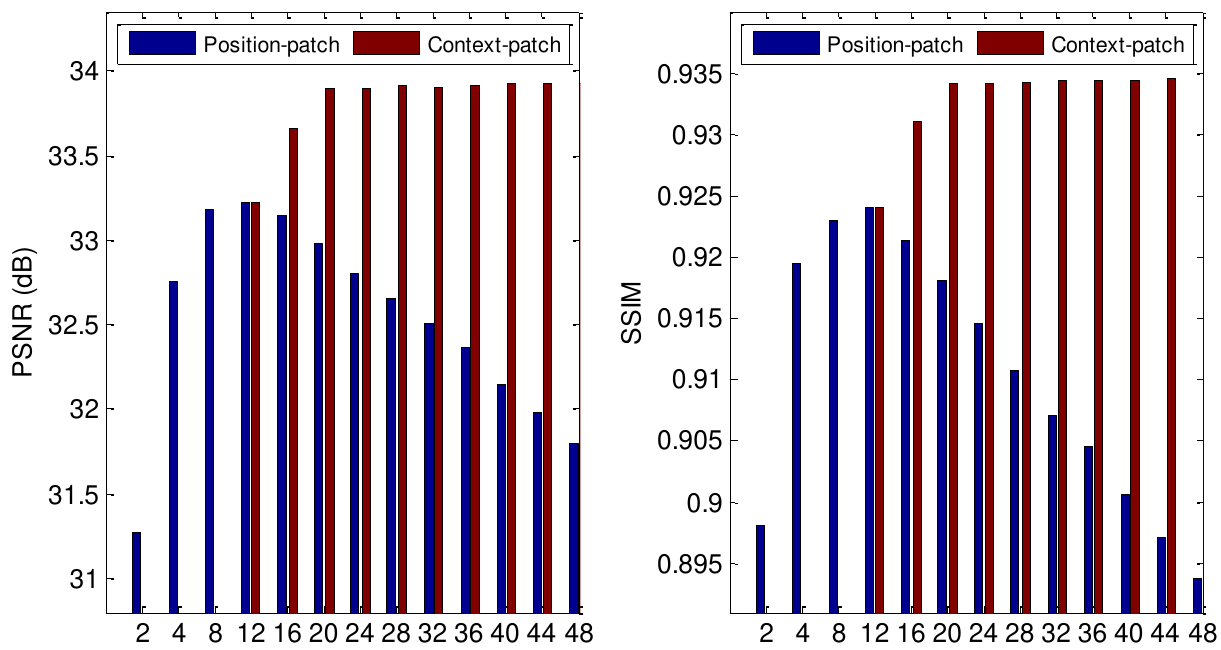}}
\vspace{-0.20cm}
\caption{Objective results of position-patch based (blue bars) and context-patch based (dark red bars) methods according to different patch and window sizes, respectively.}
\label{fig:Position_Context_PSNR}
\end{figure}


\setlength{\tabcolsep}{2.00pt}
\begin{table}[t]
\centering
 \caption{\label{tab:threestrategies}{Objective results in terms of average PSNR (dB) and SSIM of three different patch representation strategies: all-patch (using all the patches in a face image), position-patch, and context-patch.}}
 \begin{tabular}{|c||c|c|c|}
\hline
Method	&	All-patch	&	Position-patch	&	Context-patch	\\	
\hline
\hline	
PSNR	&	32.87	&	33.22	&	\textbf{33.86}	\\	
SSIM	&	0.9230	&	0.9256	&	\textbf{0.9336}	\\		
\hline
\multirow{2}{*}{Gain} &0.99 &0.64&--\\
 &0.0106&0.0080&--\\
 \hline
 \end{tabular}
\end{table}

\begin{figure}[t]
\centering
\centerline{\includegraphics[width=8.5cm]{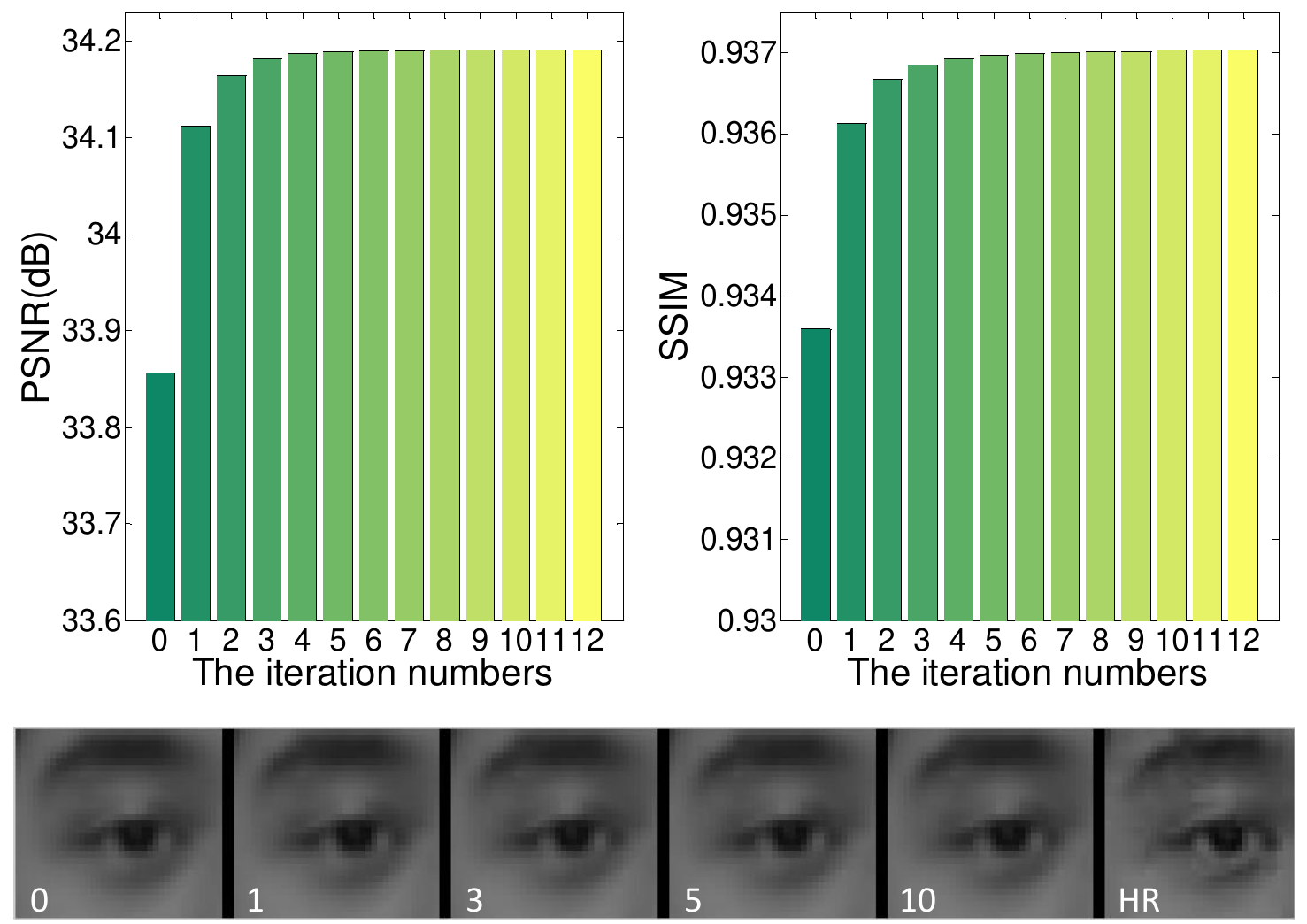}}
\vspace{-0.20cm}
\caption{Top: Plots of the objective results of the proposed TLcR-RL based face hallucination method with the increase of the iteration number in reproducing learning. Bottom: Hallucinated eye regions with different iterations and the original HR region for comparison.}
\label{fig:Iteration_PSNR}
\end{figure}

\subsubsection{Effectiveness of the locality constraint}
In Fig.~\ref{fig:paraTau} we report the average PSNR and SSIM of TLcR-RL based face hallucination method according to $\tau$. It can be seen that the performance of the algorithm rises first and then decreases. A too small or too large value of $\tau$ is not the optimal choice. When we set $\tau$ to [0.001, 0.1], the proposed TLcR-RL method will have stable improvements. Unless otherwise stated, $\tau$ is set to 0.04 in the all our experiences.

\subsubsection{Effectiveness of the hard thresholding scheme}
\label{sec:thresholding}
Fig.~\ref{fig:paraK} shows the average PSNR and SSIM according to $K$. Here, the total number of TCPs is 9000, which is calculated by substituting the patch size ($ps=12$), window size ($ss=20$), and step size ($ss=2$) to Eq.~(\ref{eq:number}), respectively. When the thresholding parameter $K$ is between 180 and 1000, the proposed TLcR-RL method continually obtains a stable and good result. When the number of $K$ is set too small, the limited training patches could not well reconstruct the input LR patch. In contrast, if the number of $K$ is set too large, it will lead to the uncertainty of representation and increase the difficulty of representing a testing patch. Unless otherwise stated, $K$ is set to 360 in the all our experiences.

\subsubsection{Effectiveness of contextual patch information}
When the window size is the same as the patch size, the proposed context-patch face hallucination method reduce to the position-patch based method. Therefore, to deomonstrate the effectiveness incorporating contextual information, we enlarge the window size to test the face hallucination performance. In Fig.~\ref{fig:Position_Context_PSNR}, the blue and dark red bars show the objective performance of position-patch based and context-patch based methods according to the patch size and window size, respectively. When the patch size is with 12$\times$12 pixels, position-patch based based method achieves the best performance. As we know, larger patch size can cover and convey more contextual information. However, when the training sample size is fixed, larger patch size does not mean better performance. This is mainly because the meaningful patch representation with large patch size calls for exponentially expanding the training set. By incorporating the contextual information, the performance of our proposed TLcR-RL based face hallucination method has a significant improvement, and reaches the stable performance at the window size of $20\times20$ pixels. When the window size is larger than 20$\times$20 pixels, there is a very slight increase. This is reasonable because when the window size is too large, the extracted context-patches will be dissimilar to the input LR patch, and will be excluded by our proposed thresholding algorithm. To balance the computational complexity and face hallucination performance, we set the window size to 20$\times$20 pixels in all our experiments.

Additionally, we consider an extreme situation when the window is set to the size of HR image, which means that we use the {all} patches on each face of the training set to represent the testing patch, the hallucination performance has a significant decrease when compared to the situation where the window size is 20$\times$20 pixels or the {position-patch method. Table \ref{tab:threestrategies} tabulates the objective results in terms of PSNR and SSIM of three different patch representation strategies, which shows again that the proposed context-patch based method is much more effective that position-patch based and all-patch based approaches. The portion-patch based method is the worst, and we attribute this to its unstable solution of patch representation when incorporating too much irrelevant training patches to the input patch. In other words, when we introduce all the image patches, the solution space of patch representation is too large (which further increasing the ill-posedness of the problem) and the locality constraint is not enough to find the optimal solution.}
\begin{table}[t]
\centering
\caption{The average PSNR (dB) and SSIM performance of LcR \cite{Jiang2014LcRTMM}, TLcR and TLcR-RL methods with different training sample sizes (T.T.S.). At the last column of each block, we give the performance gain of TLcR-RL over TLcR.}
\label{tab:RLGain}
\begin{tabular}{|c||c|c|c|c||c|c|c|c|}
\hline
\multirow{2}{*}{T.S.S.}       & \multicolumn{4}{c||}{PSNR}	& \multicolumn{4}{c|}{SSIM}\\
\cline{2-9}
 &	LcR&TLcR&TLcR-RL&(Gain)&LcR&TLcR&TLcR-RL&(Gain)\\
\hline
\hline
360	&	32.76 	&	33.86 	&	34.19 	&	\textbf{0.33} 	&	0.9145 	&	0.9336 	&	0.9370 	&	\textbf{0.0034} 	\\
300	&	32.67 	&	33.78 	&	34.15 	&	\textbf{0.38} 	&	0.9131 	&	0.9326 	&	0.9364 	&	\textbf{0.0038} 	\\
200	&	32.44 	&	33.58 	&	33.97 	&	\textbf{0.39} 	&	0.9090 	&	0.9305 	&	0.9347 	&	\textbf{0.0042} 	\\
100	&	31.75 	&	33.08 	&	33.53 	&	\textbf{0.45} 	&	0.8982 	&	0.9253 	&	0.9305 	&	\textbf{0.0052} 	\\
75	&	31.42 	&	32.91 	&	33.38 	&	\textbf{0.47} 	&	0.8922 	&	0.9238 	&	0.9291 	&	\textbf{0.0053} 	\\
50	&	30.61 	&	32.54 	&	33.12 	&	\textbf{0.58} 	&	0.8763 	&	0.9199 	&	0.9265 	&	\textbf{0.0066} 	\\
20	&	28.00 	&	31.54 	&	32.40 	&	\textbf{0.86} 	&	0.8186 	&	0.9085 	&	0.9189 	&	\textbf{0.0104} 	\\
10	&	25.66 	&	30.69 	&	31.87 	&	\textbf{1.18} 	&	0.7492 	&	0.8977 	&	0.9117 	&	\textbf{0.0140} 	\\
5	&	22.07 	&	29.95 	&	31.70 	&	\textbf{1.75} 	&	0.6228 	&	0.8937 	&	0.9082 	&	\textbf{0.0145} 	\\
\hline
\end{tabular}
\end{table}

\begin{figure}[t]
\centering
\centerline{\includegraphics[width=8.85cm]{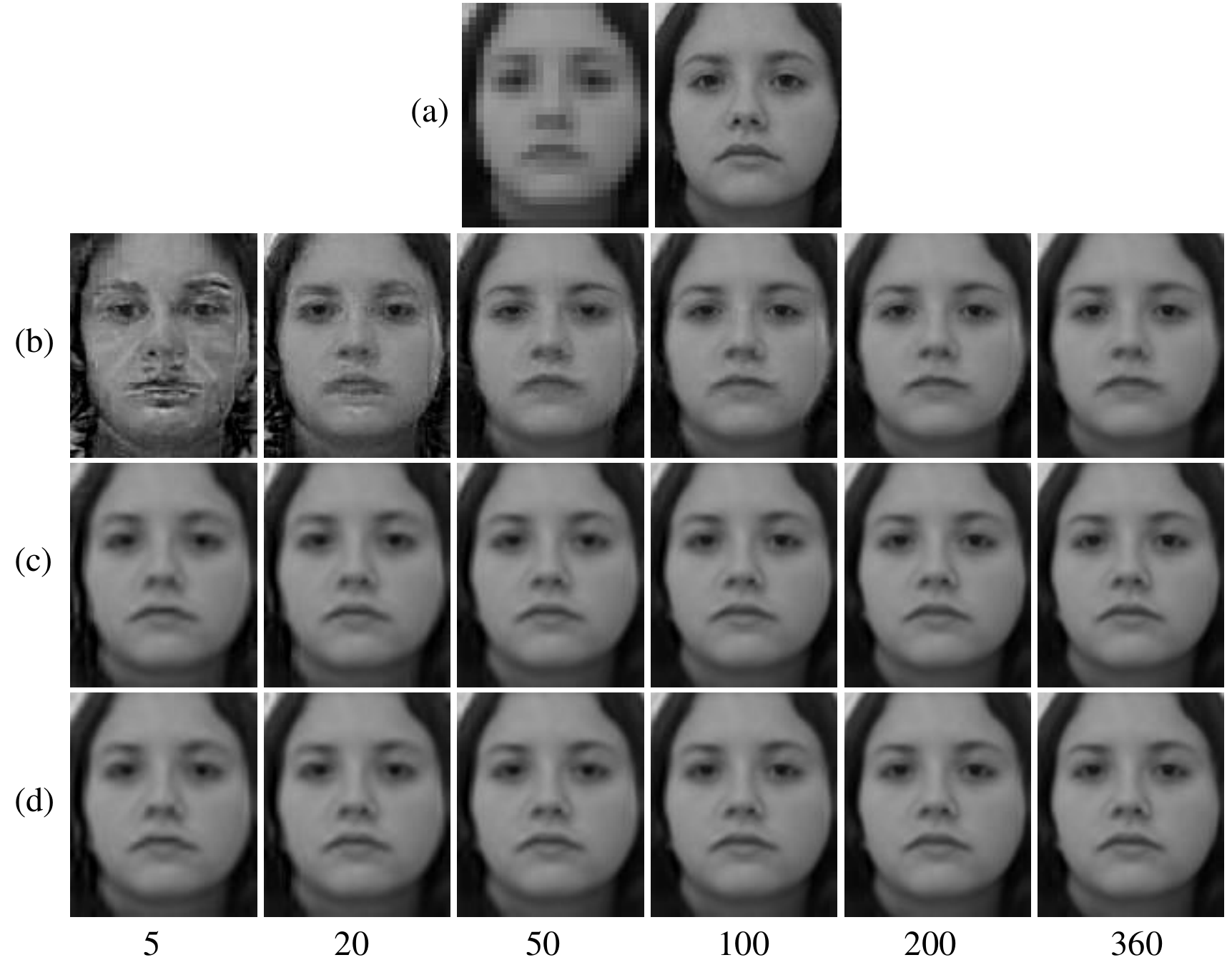}}
\vspace{-0.20cm}
\caption{Visual face hallucination results of LcR, TLcR and TLcR-RL with different training sample sizes. (a) is the input LR face and ground truth. (b), (c) and (d) are the results of LcR, TLcR and TLcR-RL methods, respectively. The numbers under the last row indicate the training sample sizes.}
\label{fig:sss}
\end{figure}

\begin{figure}[!htbp]
\centering
\centerline{\includegraphics[width=8.00cm]{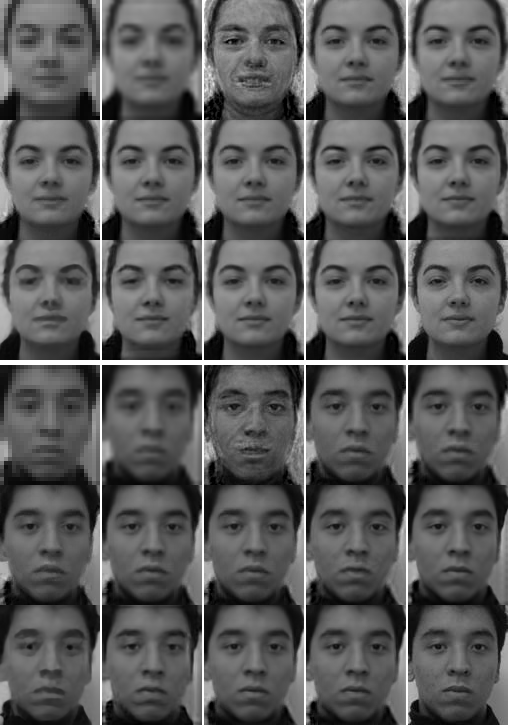}}
\vspace{-0.20cm}
\caption{Visual comparisons of two groups of hallucinated results by different face hallucination methods. For each group (every three rows), from left to right and top to bottom: the LR testing faces, hallucinated results of Bicubic interpolation method, Wang \emph{et al.}'s method~\cite{Wang2005Eig}, NE~\cite{Chang2004NE}, LSR~\cite{Ma2010LSR}, SR~\cite{Yang2010TIP}, LcR \cite{Jiang2014LcRTMM}, LINE \cite{Jiang2014TIP}, {DRP \cite{shi2015kernel}}, LCDLRR~\cite{gao2016locality}, SCN~\cite{wang2015deep}, SRCNN~\cite{dong2016image}, TLcR, and TLcR-RL. The last column is the HR ground truth.} 
\label{fig:results}
\end{figure}

\subsubsection{Effectiveness of reproducing learning}
When the HR version of the testing LR face is absent {or is dissimilar to samples in the training set}, these learning based methods will work not very well to reconstruct the ``out-of-samples''. We propose a novel improvement strategy named reproduce learning. It iteratively perform the face hallucination and training set emendation (by adding the estimated HR face to the training set). Fig.~\ref{fig:Iteration_PSNR} shows the performance according to the iteration numbers. When the iteration number is 0, it means that no estimated HR face is added to the training set. By one time reproducing learning, it has a performance improvement of 0.25 dB in term of PSNR. As the number of iterations increases, the gain becomes more and more significant. At the bottom of Fig.~\ref{fig:Iteration_PSNR}, we show the hallucinated eye regions of one LR test face with different iterations. As shown, TLcR gives the smoother result, while TLcR with reproducing learning can recover most of the detailed feature. With the increase of the iterations, the hallucinated HR eye images are much more similar to the original eye region. It reaches a stable performance after a few iterations, \emph{e.g.}, five time, which indicates the proposed method has a quick convergence.

\begin{figure*}[!htpb]
\centering
\centerline{\includegraphics[width=14.20cm]{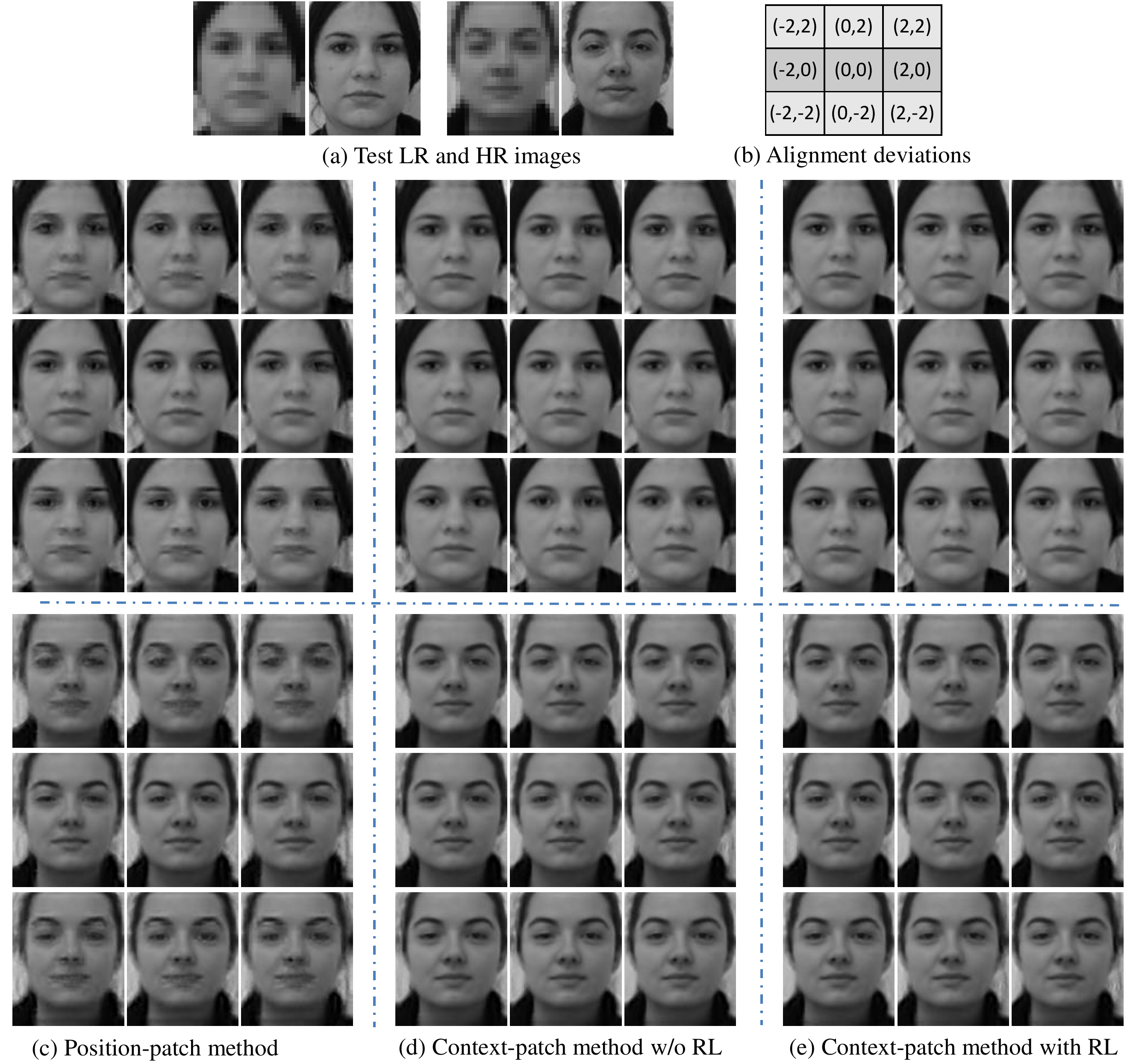}}
\vspace{-0.25cm}
\caption{Two groups of {visual} face hallucination results of position-patch and context-patch based methods with misalignment LR input. (a) are two group of test LR and HR images. (b) denotes the alignment deviations of the test LR face in pixel. (c) is the hallucinated results by position-patch method, while (d) and (e) are the results of and context-patch based methods without reproducing learning and with reproducing learning, respectively.}
\label{fig:Misalignment}
\end{figure*}

To further testify the effectiveness of the reproducing learning, we give the performance gains of TLcR-RL over TLcR when they are confronted with the SSS problem, in which case it is more likely that no similar samples can be found in the training set. In Table \ref{tab:RLGain}, we give the PSNR and SSIM performance of TLcR-RL and TLcR with different training sample sizes. Note that the results of LcR method\cite{Jiang2014LcRTMM} are also given as baselines for better comparison. With the decrease of the training sample size, the performance gains of TLcR-RL over TLcR become more and more obvious. In the very extreme situation, \emph{i.e.}, there are only five training samples in the training set, the performance gains of TLcR-RL over TLcR reach 1.75 dB (in term of PSNR) and 0.0145 (in term of SSIM). When compared with LcR~\cite{Jiang2014LcRTMM}, which is the most competitive position-patch based method, TLcR-RL is 9.63 dB better than LcR~\cite{Jiang2014LcRTMM}. Fig.~\ref{fig:sss} shows the reconstructed HR images under several typical training sample sizes. Even in the case of only five training samples, the proposed TLcR-RL still performs very well, and its hallucinated face is much clearer than that of TLcR. In contrast, when the training sample size is less than 100, it is difficult for LcR~\cite{Jiang2014LcRTMM} to reconstruct a pleasant result.

\subsection{Comparison Results}
In this section, we compare the proposed TLcR-RL with some state-of-the-arts, including Wang \emph{et al.}'s global face method~\cite{Wang2005Eig}, NE~\cite{Chang2004NE}, LSR~\cite{Ma2010LSR}, SR~\cite{Yang2010TIP}, LcR~\cite{Jiang2014LcRTMM}, LINE~\cite{Jiang2014TIP}, LCDLRR~\cite{gao2016locality}, {Shi et al.'s Dual Regularization Priors (DRP) based method \cite{shi2015kernel}}, SCN~\cite{wang2015deep} and SRCNN~\cite{dong2016image}. In addition, results of the Bicubic interpolation are the baselines. Note that NE~\cite{Chang2004NE}, SCN~\cite{wang2015deep}, and SRCNN~\cite{dong2016image} are proposed for general image reconstruction instead of face reconstruction. In the experiments, we evaluate the NE~\cite{Chang2004NE} and SRCNN~\cite{dong2016image} models by the 360 HR and LR training pairs described in Section~\ref{sec:Parameterselection}. As for SCN~\cite{wang2015deep}, we use the trained models by the authors directly to test its performance on face images. We carefully tune the parameter settings for other competitive methods to obtain their optimal performances. As for~\cite{Wang2005Eig}, the variance accumulation contribution rate is set to 99\%. In NE~\cite{Chang2004NE}, neighbor number is set to 75. We set error tolerance to 1.0 for SR~\cite{Yang2010TIP}. In LSR~\cite{Ma2010LSR} and LcR~\cite{Jiang2014LcRTMM}, the regularization parameters are set to 1e-6 and 0.04, respectively. In LINE~\cite{Jiang2014TIP}, the neighbor number, the locality regularization parameter, and the iteration number are set to 100,  1e-4, and 5, respectively. As for SRCNN~\cite{dong2016image}, we use the same image degradation as in previous methods, and the parameters are learned by the deep model. {The results of LCDLRR \cite{gao2016locality} and DRP \cite{shi2015kernel} are provided by the corresponding authors.} 

\begin{table}[t]
\centering
 \caption{\label{tab:tables}The objective results in terms of average PSNR (dB) and SSIM of different methods. The best and second best results are marked in \textcolor{red}{red} and \textcolor{blue}{blue}, respectively.}
 \begin{tabular}{|c||c|c|}
\hline
 Methods&PSNR&SSIM\\
\hline
\hline
Bicubic	&	27.50	&	0.8426	\\
\hline
Wang \emph{et al}.~\cite{Wang2005Eig}	&	27.57	&	0.7710	\\
NE~\cite{Chang2004NE}	&	32.55	&	0.9104	\\
LSR~\cite{Ma2010LSR}	&	31.90	&	0.9032	\\
SR~\cite{Yang2010TIP}	&	32.11	&	0.9048	\\
LcR~\cite{Jiang2014LcRTMM}	&	32.76	&	0.9145	\\
LINE~\cite{Jiang2014TIP}	&	32.98	&	0.9176	\\
LCDLRR~\cite{gao2016locality}	&	\textcolor{blue}{33.14}	&	0.9206	\\
DRP \cite{shi2015kernel}    & 32.84 &       \textcolor{blue}{0.9292} \\
\hline
SCN~\cite{wang2015deep}	&	32.05	&	0.9048	\\
SRCNN~\cite{dong2016image}	&	33.13	&	0.9188	\\
\hline
Our TLcR	&	33.86	&	0.9336	\\
Our TLcR-RL	&	\textcolor{red}{34.19}	&	\textcolor{red}{0.9370} \\
\hline
\hline
(\textbf{Gain})&\textbf{1.05}&\textbf{0.0078}\\
\hline
 \end{tabular}
\end{table}

In Table~\ref{tab:tables}, we give the PSNR and SSIM performance of different face hallucination methods. We observe that TLcR-RL improves the objective performance \emph{e.g.}, 1.05 dB and 0.0164 better (in terms of PSNR and SSIM) than the second best method, \emph{i.e.}, LCDLRR~\cite{gao2016locality}. We also compare with two deep learning methods, SCN~\cite{wang2015deep} and SRCNN~\cite{dong2016image}, the performance gain of our method over these two methods are still considerable. It should be noted that SRCNN~\cite{dong2016image} is retrained on the FEI database, so it can obtain better performance than SCN~\cite{wang2015deep} that used the model trained by general images. 

Fig.~\ref{fig:results} shows qualitative comparisons of TLcR-RL and other approaches on four testing images. From the visual results, we see that PCA based global face method~\cite{Wang2005Eig} has serious ghosting effects, and {its} results are dissimilar to the ground truth. The HR predictions of NE~\cite{Chang2004NE}, LSR~\cite{Ma2010LSR} and SR~\cite{Yang2010TIP} are better than Wang \emph{et al.}'s method~\cite{Wang2005Eig}, but have obvious artifacts around eyes and mouths. LcR~\cite{Jiang2014LcRTMM}, LINE \cite{Jiang2014TIP}, LCDLRR~\cite{gao2016locality}, DPR \cite{shi2015kernel}, and recently proposed deep learning methods, SCN~\cite{wang2015deep} and SRCNN~\cite{dong2016image}, are excellent methods, which can produce reasonable results that are similar to the ground truth. By carefully observing the contours of the face, eyes and nose, it can be seen that the resultant HR face images of TLcR-RL are more enjoyable and more similar to the original HR face images. In summary, the proposed TLcR-RL method demonstrates powerful hallucination ability quantitatively and qualitatively.

\begin{figure}
\centering
\centerline{\includegraphics[width=8.70cm]{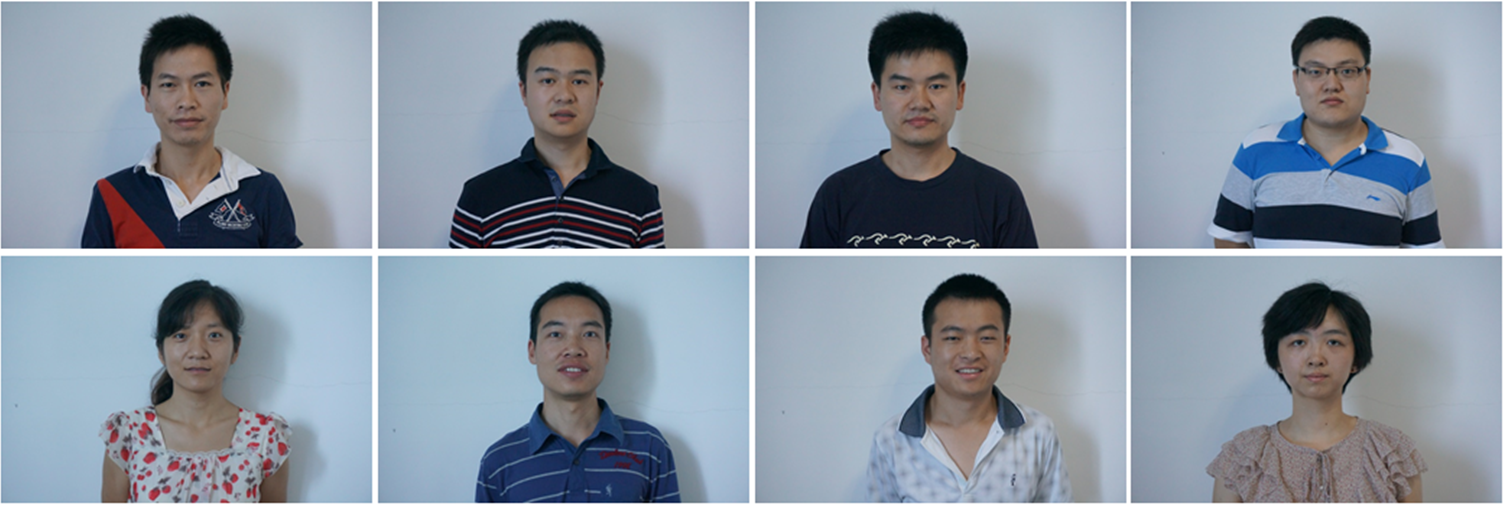}}
\vspace{-0.25cm}
\caption{Eight real-world images used to test the performance of different face hallucination methods. These images captured by an HD camera in the normal night condition. The eight face images are indexed as Img1 to Img8 in the following.}
\label{fig:HDimages}
\end{figure}

\subsection{Robustness to Misalignment}
To explore the contextual information, the context-patch based method has to introduce diverse training face image. Therefore, it will become very difficult for these position-patch based methods to find the true neighbors, \emph{i.e.}, similar training patches, for the input testing patch, especially in condition that the input face is not well aligned to the training samples. In order to demonstrate this, we conduct one subjective experiment when the observed face image is not well aligned to faces in the training set. In  Fig.~\ref{fig:Misalignment}, (a) is two well aligned faces, (b) is the alignment deviations of the test LR face in pixel, (c) shows the results of position-patch based method with different alignment deviations (b). Fig.~\ref{fig:Misalignment}(d) and (e) are the results of and context-patch based methods without reproducing learning and with reproducing learning, respectively. When the input LR face is well aligned to the training samples, both methods can well construct the {target HR images}. However, when the test face image has different alignment deviations (see Fig.\ref{fig:Misalignment} (b), (0,0) indicates the observed face is well aligned to the faces in the training set), the reconstructed HR face of position-patch method has obvious ghosting effects. In contrast, the proposed context-patch based method can produce clear and shape edges. When compared with the hallucinated results with and without reproducing learning, we can see that the latter can well capture the facial details (please refer to the eye regions of these two methods). This once again proves the validity of the proposed reproducing learning algorithm.

\begin{figure*}
\centering
\centerline{\includegraphics[width=15.20cm]{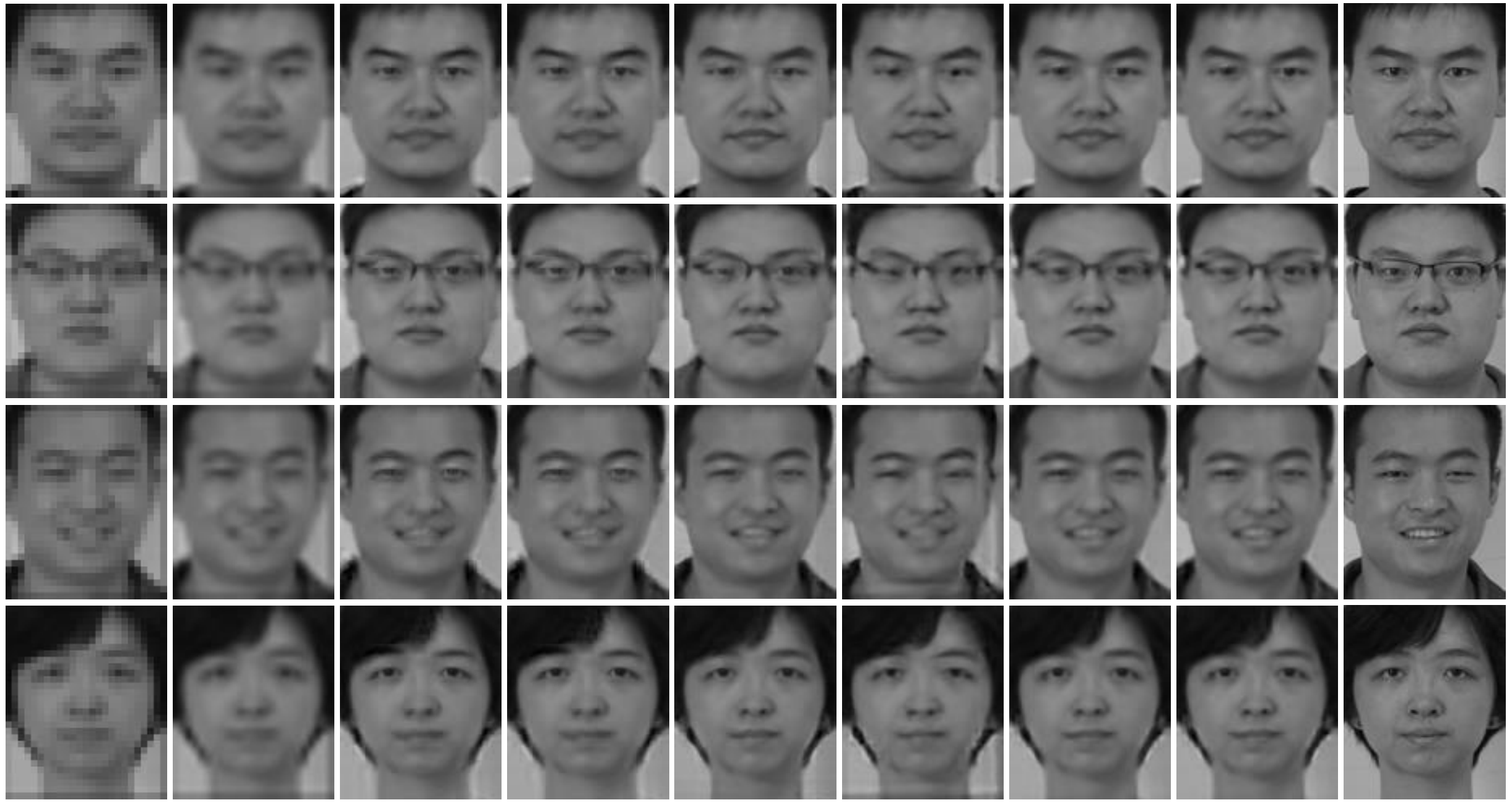}}
\vspace{-0.25cm}
\caption{Visual face hallucination results of the four real-world images: Img3, Img4, Img7, and Img8. From the first column to the last column: the LR test faces, results of Bicubic, LcR~\cite{Jiang2014LcRTMM}, LINE~\cite{Jiang2014TIP}, {DRP \cite{shi2015kernel}}, SRCNN~\cite{dong2016image}, TLcR and TLcR-RL, the ground truth HR faces.}
\label{fig:Results8}
\end{figure*}

\subsection{Hallucinating with Real-World Images}
\label{sec:realworld}
In this subsection, we conduct one another experiment to demonstrate the effectiveness and the advancement of the proposed algorithm with some real-world face images that are very different from the face image in FEI database. We capture eight high-definition images as shown in Fig.~\ref{fig:HDimages}.

\begin{figure}[h]
\centering
\centerline{\includegraphics[width=8.80cm]{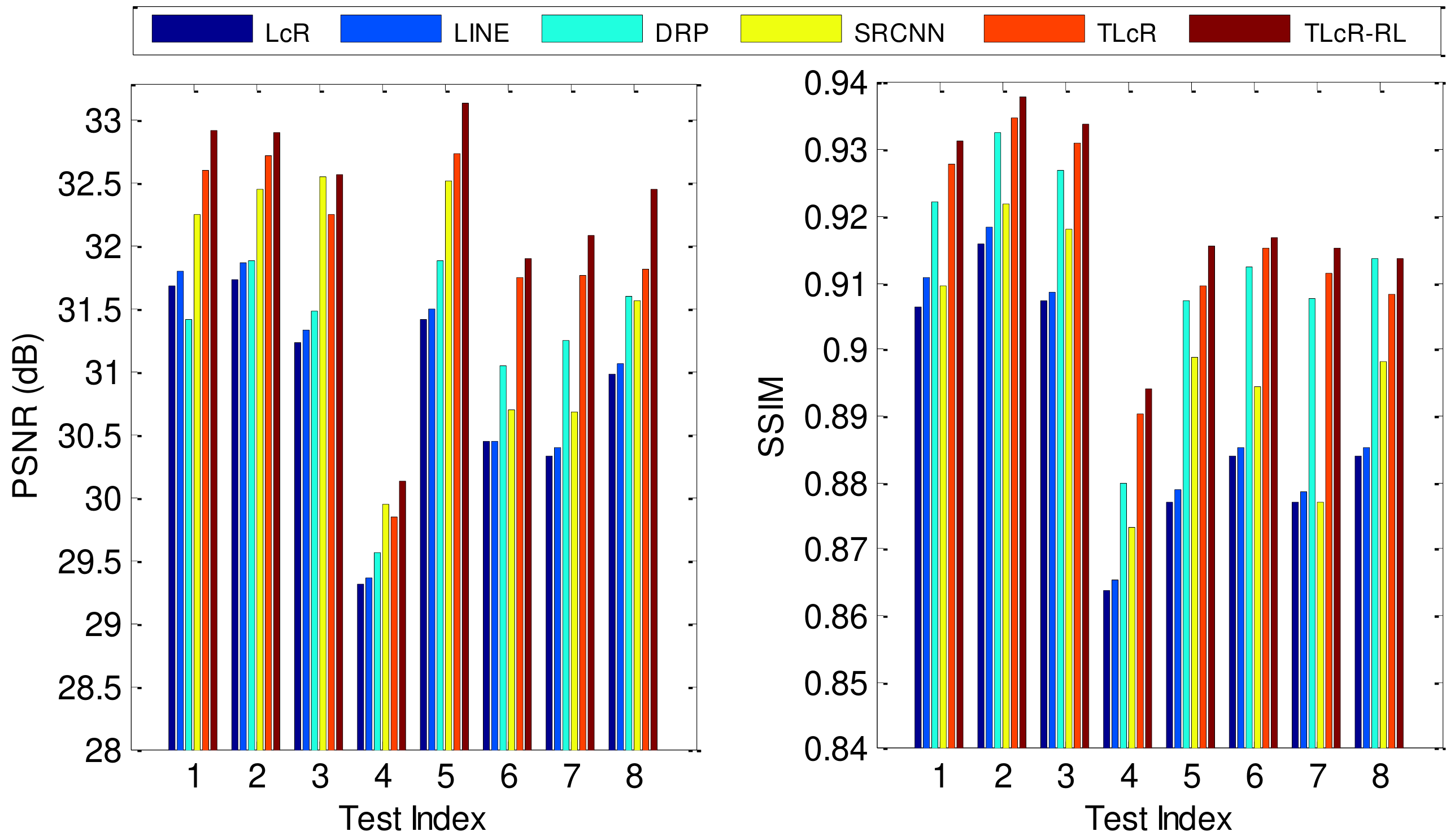}}
\vspace{-0.25cm}
\caption{PSNR and SSIM results of five competitive methods on the eight testing images. The average PSNR of these methods are 30.90 dB, 30.98 dB, 31.58 dB, 31.94 dB, and 32.26 dB, while the average SSIM of these methods are 0.8895, 0.8915, 0.8989, 0.9160, and 0.9198.}
\label{fig:Imgs8_PNSR_SSIM}
\end{figure}

Firstly, the commonly used automatic face detection algorithm~\cite{Sivic2006BuffyFaceRecognition} is applied to detect the face regions in the captured HR images, and then we align the detected faces to the mean face by the detected two points of eye centers\footnote{The automatic face detection algorithm outputs the coordinates of eye corner. We then use the mean coordinates to represent the eye center.}. Then, they are cropped to 120$\times$100 pixels, which are the ground truth HR faces (see the last column of Fig.~\ref{fig:Results8}). In our experiments, the LR test faces are obtained by the same way as in Section~\ref{sec:Parameterselection} (see the first column of Fig.~\ref{fig:Results8}). Here we only compare with two most representative local position-patch based methods, \emph{e.g.}, LcR \cite{Jiang2014LcRTMM} and LINE~\cite{Jiang2014LcRTMM}, one {global reconstruction constraint patch-based method, DRP \cite{shi2015kernel}}, and the deep learning based method, \emph{e.g.}, SRCNN~\cite{dong2016image}, for their representative and good performance. The middle seven columns are the hallucinated results by Bicubic, LcR~\cite{Jiang2014LcRTMM}, LINE \cite{Jiang2014TIP}, SRCNN~\cite{dong2016image}, TLcR and TLcR-RL. Note that, for these color face images, we change the input face images from RGB space to YUV space firstly, and then reconstruct them in the luminance component. {This is mainly because humans} are more sensitive to illuminance changes. The hallucinated eyes and face contours by SRCNN~\cite{dong2016image} are blurry and dirty. LcR \cite{Jiang2014LcRTMM} and LINE \cite{Jiang2014TIP} produce some ghosting effects around the eyes, mouths and face contours. {DRP \cite{shi2015kernel} and the proposed can well maintain the face contours.} Fig.~\ref{fig:Imgs8_PNSR_SSIM} plots the results of these methods on the eight testing images. The proposed methods also show the best objective results. By further examination, it can be seen that the improvement on SSIM is much more obvious than that on PSNR, which indicates that our face hallucination model pays more attention to the face structure information and the hallucinated results are much more consistent with visual perception. From the PSNR results of Img3 and Img4, we find that TLcR is worse than SRCNN~\cite{dong2016image}. However, the hallucinated results (see the first two rows of Fig.~\ref{fig:Results8}) of TLcR is much better than that of SRCNN~\cite{dong2016image}. This observation is consistent with the SSIM results. {The SSIM results of DRP \cite{shi2015kernel} are very competitive and this demonstrates its ability in maintaining the face structures.} The above experiments indicate the effectiveness of the proposed TLcR-RL method in the real-world condition.

\section{Main Findings and Future Work}
\label{sec:conclusions}
In this paper we introduce a context-patch based face hallucination approach that can fully exploit contextual information of image patches. Different from conventional position-patch based approaches, which use only the training patches from the same position as the testing patch for reconstruction, the proposed method leverages the contextual information and uses the TCPs to obtain a better representation. In order to improve the accuracy of reconstruction, we have developed a hard threshold scheme to avoid being affected by these dissimilar training patches. In addition, we have also developed an iterative enhancement strategy to improve the estimation results by reproducing learning. The experiment verifies its robustness to misalignment and the SSS problem. Comparison results with some competitive approaches, including two deep learning based super-resolution methods, show that the hallucinated face images of our proposed approach have finer and more detailed features over state-of-the-arts.

In TLcR-RL, the representation coefficients of the input sample can be seen as the filter responses by a set of filters (or basis, \emph{i.e.}, the training samples in our method), while the thresholding can be seen as the output of an activation function. The overlapped patch averaging {strategy} can be regarded as the filtering on a set of feature maps by some pre-defined filters. The above three operations can be formed as a convolutional layer. By incorporating reproducing learning, the proposed TLcR-RL is very similar to the form of DNNs. Therefore, how to combine the structure prior and the very efficient deep learning algorithms will be our first concern in the future.

In our experiments, we mainly focus on the reconstruction with frontal portrait in well controlled conditions, how to extend the proposed model to uncontrolled conditions, such as variety in poses, expression and illumination, will be our second future work.

\section*{Acknowledgment}
We would like to thank Dr. Gao, the first author of \cite{gao2016locality}, for his kind providing of the results of LCDLRR algorithm. We would like to thank the authors of~\cite{Ma2010LSR}, \cite{Yang2010TIP}, \cite{wang2015deep}, \cite{dong2016image}, and DRP \cite{shi2015kernel} for their kind sharing of their source codes.


%

%


\ifCLASSOPTIONcaptionsoff
  \newpage
\fi



%
{
\footnotesize
\bibliographystyle{IEEEtran}
\bibliography{TIP2017}
}




\end{document}